\documentclass[aps,preprint,groupedaddress]{revtex4}%PRE
\usepackage{amsfonts}
\usepackage{mathrsfs}
\usepackage{graphicx}
\usepackage{amsmath}
\usepackage{amssymb}
\usepackage{amsbsy}
\usepackage{bm}
\usepackage{soul}
\usepackage{xcolor}

\begin{document}

\title{Data-driven effective model shows a liquid-like deep learning}
\author{Wenxuan Zou}
\author{Haiping Huang}
%\email{huanghp7@mail.sysu.edu.cn}
\affiliation{PMI Lab, School of Physics,
Sun Yat-sen University, Guangzhou 510275, People's Republic of China}

\date{\today}
\begin{abstract}
The geometric structure of an optimization landscape is argued to be fundamentally important to support the success of deep neural network learning. 
A direct computation of the landscape beyond two layers is hard. Therefore, to capture the global view of the landscape,
an interpretable model of the network-parameter (or weight) space must be established. However, the model is
lacking so far. Furthermore, it remains unknown what the landscape looks like for deep networks of binary synapses, which plays a key role in robust and energy efficient 
neuromorphic computation.
Here, we propose a statistical mechanics framework by directly building a least structured model of the high-dimensional weight space,
considering realistic structured data, stochastic gradient descent training, and the computational depth of neural networks.
We also consider whether the number of network parameters outnumbers the number of 
supplied training data, namely, over- or under-parametrization. Our least structured model reveals that the weight spaces of the under-parametrization and
over-parameterization cases belong to the same class, in the sense that these weight spaces are well-connected without any hierarchical clustering structure. 
In contrast, the shallow-network has 
a broken weight space, characterized by a discontinuous phase transition, thereby clarifying the benefit of depth in deep learning from the angle of
high dimensional geometry.
Our effective model also reveals that inside a deep network, there exists a liquid-like 
central part of the architecture in the sense that the weights in this part behave as randomly as possible, providing algorithmic implications. 
Our data-driven model thus provides
a statistical mechanics insight about why deep learning is unreasonably effective in terms of the high-dimensional
weight space, and how deep networks are different from shallow ones.
\end{abstract}

 \maketitle

%%%%%%%%%%%%%%%%%%%%%%%%%%%%%%%%%%%%%%%%%%%%%%%%%%%%%%%%%%%%%%%%%%

\section{Introduction}
\label{Intro}
 Artificial deep neural networks have achieved state-of-the-art performance in many industrial and academic domains ranging from
 pattern recognition and natural language
 processing~\cite{DL-2016} to many-body quantum physics and classical statistical physics~\cite{RMP-2019}. However, it remains challenging to reveal the mechanisms 
 underlying the success of deep learning. One key obstacle is building the causal relationship between the high-dimensional non-convex optimization landscape and
 the state-of-the-art performance of deep networks~\cite{Sejnowski-2020}. In the past few years, many theoretical and empirical efforts contributed to understanding this fundamental relationship
 in deep networks of real-valued weights.
 High-loss local minima of the optimization landscape are rare in the over-parametrization regime where the number of trainable parameters (weights) in a typical deep network is much larger than
 the number of training data~\cite{Geiger-2019}. All minima in the landscape were shown to be global minima, given special requirements for network architectures and neural activation functions~\cite{Nguyen-2017}.
 In particular, no substantial barriers between minima were observed in deep learning~\cite{Draxler-2018}. These minima can be even connected via hyper-paths within a unique global valley~\cite{Gari-2018,Nguyen-2019,Fort-2019b}.
 The geometric structure of local minima was also extensively studied~\cite{Ballard-2017,Sagun-2017,Li-2018,Fort-2019a}, in terms of a controversial description of the geometric flatness of the minima~\cite{Hochreiter-1997,Keskar-2016,Dinh-2017,Baldassi-2019,Tu-2021}. 
 The flat minima imply that the weights around them can be perturbed without significantly changing the training cost function.
 The flatness of such minima was empirically supported by the eigen-spectrum of the Hessian matrix evaluated at those minima~\cite{Sagun-2017}. The curvature of the landscape was also claimed to correlate with the
 generalization ability of deep networks for unseen data~\cite{Chaudhari-2016,Tu-2021}. Therefore, studying deep networks from the perspective of the
 high-dimensional weight space is fundamentally important, yet the previous studies did not give a global view about the collective behavior of interacting weights.
 
 Physics methods have been applied to investigate the weight space structure of non-convex high-dimensional problems, e.g., the spherical or binary spin model approximation of deep neural networks~\cite{Fyo-2007,Lecun-2014,Huang-2018,Becker-2020}, and
 the weight space structure of binary perceptron problems~\cite{Huang-JPA2013,Huang-2014,Baldassi-2015}. However, one drawback of these studies concerns their strong assumptions to build toy models. These models assume unstructured data as well as
 shallow networks, and moreover the learning behavior is far from the practical training procedure used in a typical deep learning. Therefore, it is computationally hard to directly compute the weight space even beyond two-layer architectures, and moreover
 it remains elusive from an interpretable-model perspective 
 what the geometric structure of the deep-learning landscape looks like and which key factor affects the landscape. In this paper, we restrict the theoretical analysis to deep networks of binary synapses, which is particularly important 
 for robust and energy-efficient neuromorphic computation~\cite{Esser-2016,Roy-2019}.
 
 Here, from a conceptually different viewpoint, we propose a data-driven effective model, taking into account interactions among structured data, deep architectures, and training.
 We consider deep networks of binary weights that are amenable for theoretical analysis and moreover can be efficiently trained with mean-field methods~\cite{Shayer-2018}. The weight configurations realizing a
 high classification accuracy of the structured data are collected as
 a large ensemble, whose statistics is described by first-order and second-order moments of the weight distribution.
 To construct an effective model of the weight space,
 the maximum-entropy principle~\cite{Jaynes-1957} is applied. This principle was previously used 
 to analyze neural population at the collective network-activity level~\cite{Nature-2006,Huang-2016}. 
 This effective model of the practical deep learning is 
 then analyzed from an entropy landscape angle, in which the geometric structure of the entire weight space 
 can be characterized in different contexts: over-parametrization, under-parametrization and shallow networks.
 
Our study establishes an interpretable model of the weight space focusing on
the global view of the landscape, and this model
has three important predictions: (i) The deep-learning weight-space is smooth and dominated by a single well-connected component, 
while the shallow network is characterized by a broken weight space. (ii) For the deep network, the
under-parameterization and over-parameterization (the number of training data is more and less than the number of parameters, respectively) 
belong to the same universal class with a common smooth landscape, which exactly coincides with recent empirical findings of no substantial 
barriers between minima in the deep-learning loss landscape~\cite{Gari-2018,Draxler-2018,Nguyen-2019}, although we focus on the deep networks of binary weights.
We remark that the universal class may be related to the absence of the double descent phenomenon for the deep networks of binary weights, at least in the explored range of training parameters.
(iii) A most surprising prediction is that a special interior part of a largest entropy in the weight space emerges after learning, 
showing a liquid-like property, compared to two more constrained boundaries of the deep network. This part is conjectured to play a key role in fast learning dynamics.
%%%%%%%%%%%%%%%%%%%%%%%%%%%%%%%%%%%%%%%%%%%%%%%%%%%%%%%%%%%%%%%%%%%%
\begin{figure*}
	\centering
	\includegraphics[bb=26 12 2356 653,width=\textwidth]{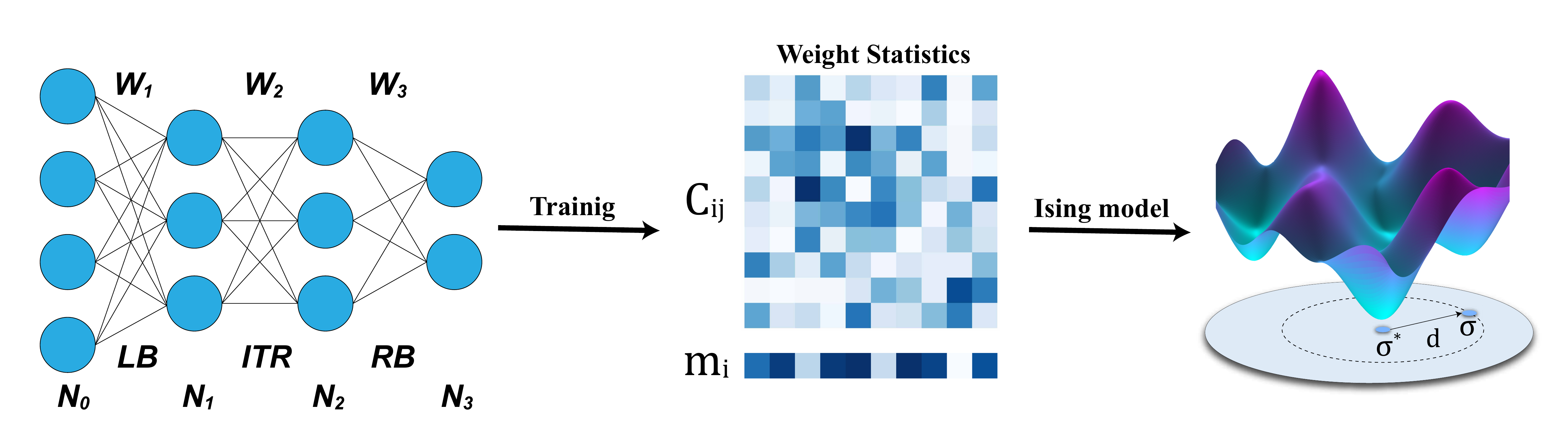}
	\caption{Schematic illustration of how to construct an effective physics model analyzing the deep-learning weight space.
We consider a four-layer neural network of binary weights to perform a classification task, 
where the input data pass through the weight regions of the left boundary (LB), 
the interior (ITR), and the right boundary (RB) in order. After the neural network is trained from 
different initializations, the weight statistics including magnetizations (first-order moments) 
$m_i = \langle \sigma_i\rangle$ and pairwise correlations (second-order moments) 
$C_{ij}=\langle\sigma_i\sigma_j\rangle$ will be computed directly from the collected weight configurations.
Based on the weight 
statistics, we establish an effective data-driven Ising model that is able to reflect 
the weight space characteristics, and then apply the entropy landscape analysis by introducing a distance-dependent 
extra term into the Boltzmann distribution of the effective Ising model. The entropy landscape analysis can help to 
explore the internal structure of the weight parameter space by counting how dense weight 
parameters $\boldsymbol{\sigma}$ are at a typical distance $d$ away from the reference one 
$\boldsymbol{\sigma}^*$.}
	\label{DNN}
\end{figure*}
\section{Deep Learning Setting}
\label{Model}
In order to collect weight configurations 
in the neural network parameter space, we design a neural network architecture to solve the 
classification task of the handwritten digits (or the MNIST dataset~\cite{mnist}) first, and then an extensive number of weight solutions that lead to high test accuracy are collected.
The test accuracy refers to the network's ability to classify unseen dataset. More precisely,
we consider a four-layer fully-connected feedforward network. Each layer has $N_l$ $(l = 0,1,2,3)$ neurons,
where $N_0$ equals to the dimension of the input vector, and $N_3$ equals to the number of total classes. 
The weight matrix $\mathbf{W}^{l}$ indicates connections between layer $l$ and layer $l-1$ (Fig.~\ref{DNN}). For simplicity, we 
do not consider biases for neurons. The neural activity (including the input) is transfered by a non-linear function, i.e., the leaky ReLU (L-ReLU)
function defined by $\max(0,z)+0.1\min(0,z)$, where $z$ denotes the pre-activation or the weighted-sum of input activities.
We finally add a softmax layer after the output layer and use the cross entropy as the loss function. 

To collect a sufficient number of weight configurations and facilitate theoretical analysis, we require that the
neural network architecture should meet two demands. First, the model complexity, 
namely, the number of the total weights should be small. This avoids a sharp growth of the computational cost consumed to collect a huge number of weight configurations.
Second, the weight of the neural network should take binary values $\{+1,-1\}$, as 
the following analyses are based on the Ising model, in which the weights are analogous to the spins in a physics model. 

To fulfill the first demand, we should first create a set of input data with low dimension. 
Thus, we carry out a principal component analysis (PCA) of the MNIST dataset, retaining only the top $20$ principal components, 
which explain about $64.5\%$ of the total variance in the original data. After this dimension reduction, a relatively high-quality reconstruction of the digits can be performed.
Although the loss of information will degrade the test accuracy, we can still reach a relatively high test accuracy 
(see Appendix~\ref{app-c}).

Because of the second demand, the network with binary weights is difficult 
to train by the standard gradient decent algorithm, due to the fact that the cost function is not differentiable with respect to
the discrete synaptic weight. 
To tackle this challenge, we apply a recently-proposed mean-field training algorithm~\cite{Shayer-2018}. More precisely, 
we consider the situation that each weight $w^l_{ij}$ follows a 
binomial distribution $P(w^l_{ij})$ parametrized by an external field $\theta^l_{ij}$ as 
follows:
\begin{equation}\label{eq:p}
P(w^l_{ij}) = \sigma(\theta^l_{ij})\delta_{w^l_{ij},1} + [1-\sigma(\theta^l_{ij})]\delta_{w^l_{ij},-1},
\end{equation}
where $\sigma(\cdot)$ is a sigmoid function. The external fields are clearly continuous. Therefore, the standard backpropagation
can be applied to 
the external fields $\{\theta^l_{ij}\}$, instead of weights (ill-defined in gradients).

In the feedforward transformation, the pre-activation can be written 
as $z_j^l = \sum_i w^l_{ij} a_i^{l-1}$, where $a_i^{l-1}$ is the activation at the layer $l-1$. 
The weight $w^l_{ij}$ is subject to the aforementioned binomial distribution, and thus its mean and variance, i.e., $\mu^l_{ij}$ 
and $(\sigma^l_{ij})^2$, are given by:
\begin{subequations}
\begin{align}
&\mu^l_{ij}=\langle w^l_{ij}\rangle=2\sigma(\theta^l_{ij})-1,\\
&(\sigma^l_{ij})^2 = \langle(w^l_{ij})^2\rangle-\langle w^l_{ij}\rangle^2=-4\sigma^2(\theta^l_{ij})+4\sigma(\theta^l_{ij}).
\end{align}
\end{subequations}
To avoid a direct sampling process, we use the local re-parametrization trick~\cite{Kingma-2015}. 
According to the central limit theorem, $z_j^l$ can then be approximated to follow a 
Gaussian distribution $z_j^l \sim \mathcal N\Big(\sum_i\mu^l_{ij}a_i^{l-1},\sum_i(\sigma^l_{ij})^2 (a^{l-1}_{i})^2\Big)$. Hereafter, 
we define the mean and variance of $z_j^l$ to be $m_j^l$ and $(v_j^l)^2$, respectively. Taken together, 
the feedforward transformation can be finally re-parametrized as:
\begin{subequations}
\begin{align}
&z_j^l = m_j^l + v_j^l\cdot \epsilon_j^l ,\\
&a_j^l = \frac{1}{\sqrt{N_{l-1}}} \operatorname{L-ReLU}(z_j^l),
\end{align}
\end{subequations}
where $\epsilon_j^l$ is a standard Gaussian random variable, and $\frac{1}{\sqrt{N_{l-1}}}$ is a scaling factor to eliminate 
the layer-width dependence of the activation $a_j^l$. Detailed training procedures are given to Appendix~\ref{app-a}.

\begin{figure*}
	\centering
	\includegraphics[bb=8 6 1317 412,width=\textwidth]{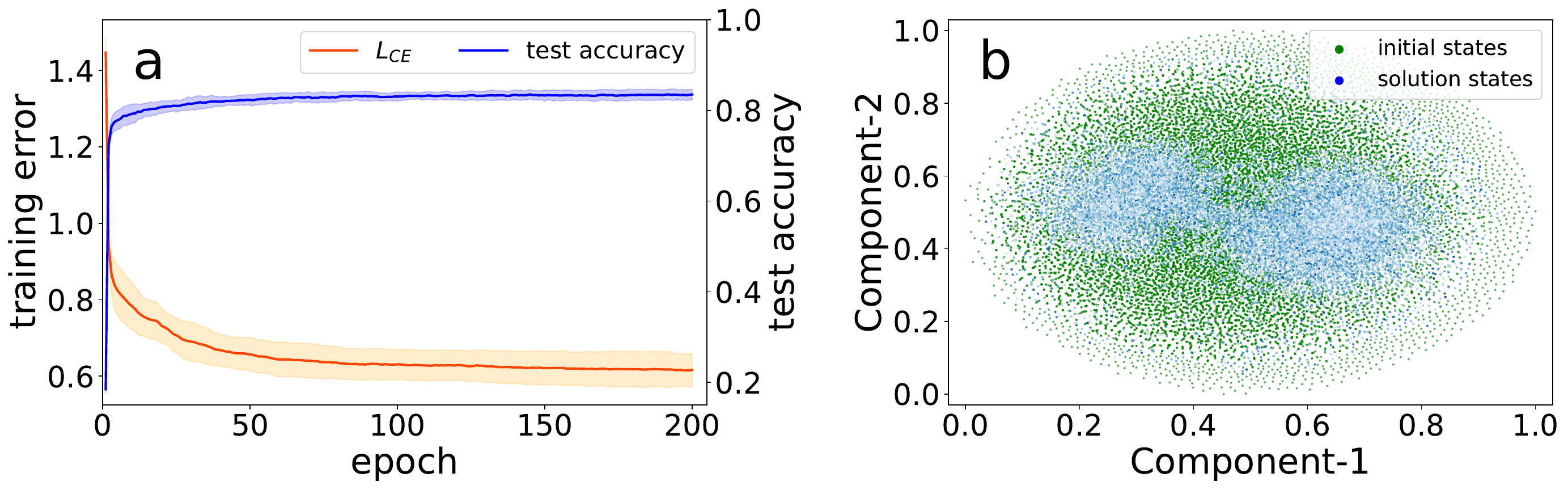}
	\caption{Neural network training and properties of weight configurations sampled from independent runs of training ($P=60\ 000$, $N=675$). 
(a) Training trajectories of four-layer networks. The cross-entropy (the orange line) 
and test accuracy (the blue line) can reach $0.61$ and $83.5\%$, respectively at the $200$-th epoch, 
where the fluctuation indicated by the shadow regime is computed from ten independent training runs. 
(b) Two-dimensional representations 
of $20\ 000$ random initializations (green dots) and $20\ 000$ learned weight configurations (blue dots). The low dimensional projection is 
carried out by the t-SNE technique~\cite{tsne-2008}.
The random initialization denotes the weight configuration (or initial state) sampled by the Gaussian 
random initial fields $\boldsymbol{\theta}_0$. The collected weight configurations (or solution states)
are obtained at the end of the 
training starting from the initial fields $\boldsymbol{\theta}_0$.
 For the solution states, the deeper their colors are, the higher the corresponding test accuracies are, 
while the random initial states are all of the same green color of chance-level accuracy. }
\label{tsne}
\end{figure*}

In practice, we use two hidden layers with $15$ neurons for each. 
The widths of input layer and output layer are determined by the data, 
which are $20$ and $10$, respectively. Thus, our network architecture is specified by 20-15-15-10, resulting in 
the number of weight parameters $N=675$. We highlight that the total number of possible weights $2^N\approx10^{1213}$ is already very huge for
a statistical analysis. As expected, the size of hidden layer makes learning improve over the shallow counterpart (see Appendix~\ref{app-g}).
Unless stated otherwise, we use the entire MNIST dataset (i.e., $60\ 000$ digits examples) for training and the other 
$10\ 000$ examples for testing. In the following analysis, we denote $P$ as the number of training examples. 
By fine tuning hyper-parameters of the training procedure (see Appendix~\ref{app-c}), the 
neural network can reach a test accuracy of $85\%$ [Fig.~\ref{tsne}(a)], i.e., the trained network can classify correctly
$85\%$ of the unseen test dataset.  
To collect distinct weight configurations, we train the neural network 
for $50$ epochs starting from a random Gaussian initialization of external fields $\boldsymbol{\theta}$ during each sampling trial. 
Then, we use the trained external fields to generate $10$ weight configurations and select the one with 
the optimal test accuracy ($\ge 80\%$). In this way, we obtain two million weight configurations (the corresponding number is specified by $M$ below)
with test accuracies ranged from $80\%$ to $83\%$, for constructing an effective model.

To give a brief picture of the weight parameter space, 
we apply the t-SNE dimension reduction technique~\cite{tsne-2008} to $20\ 000$ random initialization points and the corresponding 
learned weight configurations with the same number [Fig.~\ref{tsne} (b)]. Here, random initialization points represent the weight parameters 
sampled from the Gaussian-initialized external fields, from which the external 
fields are trained to reach the test accuracy threshold. The collected weight configurations are sampled from the final binomial distribution.
Note that the dominant part of the weight configuration space (valid weights realizing the classification task)
seems to be organized into a large connected component [Fig.~\ref{tsne} (b)], which also indicates that 
compared to the entire parameter space, the weight space realizing the classification task occupies a relatively small region.
Next, we design a semi-rigorous statistical mechanics framework to explore a detailed high-dimensional geometric structure of the weight space for 
the current deep-learning model.

\section{Data-driven Ising model}
\label{Ising}
We build a data-driven Ising model to describe these $M$ weight configurations, 
as we are interested in the statistical properties of the deep-learning weight space. 
More precisely, one weight parameter solution $\mathbf{W}=\{w_1,w_2,\ldots,w_N\}$ is
transformed to a spin configuration $\boldsymbol{\sigma}=\{\sigma_{1}, \ldots, \sigma_{N}\}$ in the following 
analysis. Given the constraints of weight statistics including magnetization (first-order moments) 
$m_i = \langle \sigma_i\rangle$ and pairwise correlation (second-order moments) 
$C_{ij}=\langle\sigma_i\sigma_j\rangle$, the least-structured-model probability distribution 
$P\left(\boldsymbol{\sigma}\right)$ to fit the weight statistics follows
the maximum entropy principle~\cite{Jaynes-1957} (see also Fig.~\ref{DNN} for an illustration). According to this principle, $P\left(\boldsymbol{\sigma}\right)$ is recast into the form of
the Boltzmann distribution: 
\begin{equation}
\label{BZD}
 P\left(\boldsymbol{\sigma}\right)=\frac{1}{Z}\exp(-\beta E(\boldsymbol{\sigma})),
\end{equation} 
where $Z$ is the partition function in statistical physics, $\beta$ is the inverse temperature ($\beta=1$ throughout the paper as it can be absorbed into the model parameters),
and $E\left(\boldsymbol{\sigma}\right)=-\sum_{i<j} J_{i j} \sigma_{i} \sigma_{j}-\sum_{i} h_{i} \sigma_{i}$ 
is the energy that is consistent with the Ising-model's Hamiltonian. The model parameter $J_{ij}$ 
denotes the functional coupling between weight $\sigma_i$ and weight $\sigma_j$ interpreting the second-order 
correlation among weights, where $J_{ij}>0$ and $J_{ij}<0$ refer to ferromagnetic 
and anti-ferromagnetic interactions, respectively. The model parameter $h_i$ denotes the bias of the weight $\sigma_i$, reflected by  
the external field on the weight $\sigma_i$ after training.

To find the model parameters $\boldsymbol{\Omega}=\{\mathbf{J},\mathbf{h}\}$, we use the
gradient accent algorithm corresponding to the maximum likelihood learning principle. 
By maximizing the log-likelihood $\langle\ln \left(P_{\boldsymbol{\Omega}}(\boldsymbol{\sigma})\right)\rangle_{\text {weight-data}}$ 
with respect to the model parameters $\boldsymbol{\Omega}=\{\mathbf{J},\mathbf{h}\}$, 
we can obtain the following iterative learning rules~\cite{Ackley-1985}:
\begin{subequations}\label{eq:learning}
\begin{align}
&J_{i j}^{t+1}=J_{i j}^{t}+\eta\left(\left\langle\sigma_{i} \sigma_{j}\right\rangle_{\text {data }}-\left\langle\sigma_{i} \sigma_{j}\right\rangle_{\text {model }}\right),\\
&h_{i}^{t+1}=h_{i}^{t}+\eta\left(\left\langle\sigma_{i}\right\rangle_{\text {data }}-\left\langle\sigma_{i}\right\rangle_{\text {model }}\right) ,
\end{align}
\end{subequations}
where $t$ and $\eta$ denote the learning step and learning rate, respectively. 
In the above learning equations, the data expectation terms can be directly 
computed from the sampled parameter solutions. However, the model expectation terms are commonly difficult to 
compute due to the $\mathcal{O}(2^N)$ computation complexity. Fortunately, the cavity method in spin glass theory (Appendix~\ref{app-b}) can be used
to approximate the model expectations. At each learning step, to evaluate how well 
the Ising model fits the weight statistics, we compute the root-mean squared error (or deviation) between the data expectations and model expectations:
$\Delta^2=\frac{1}{N} \sum_{i}\left[m_{i}^{\text {data}}-m_{i}^{\text {model}}\right]^{2}+\frac{2}{N(N-1)} \sum_{i<j}\left[C_{i j}^{\text {data}}-C_{i j}^{\text {model}}\right]^{2}$,
from which a perfect fitting leads to a zero deviation. We stop the learning when $\Delta<0.01$ or 
after $300$ learning steps are reached. 

\begin{figure*}
	\centering
	\includegraphics[bb=6 2 1829 986,width=\textwidth]{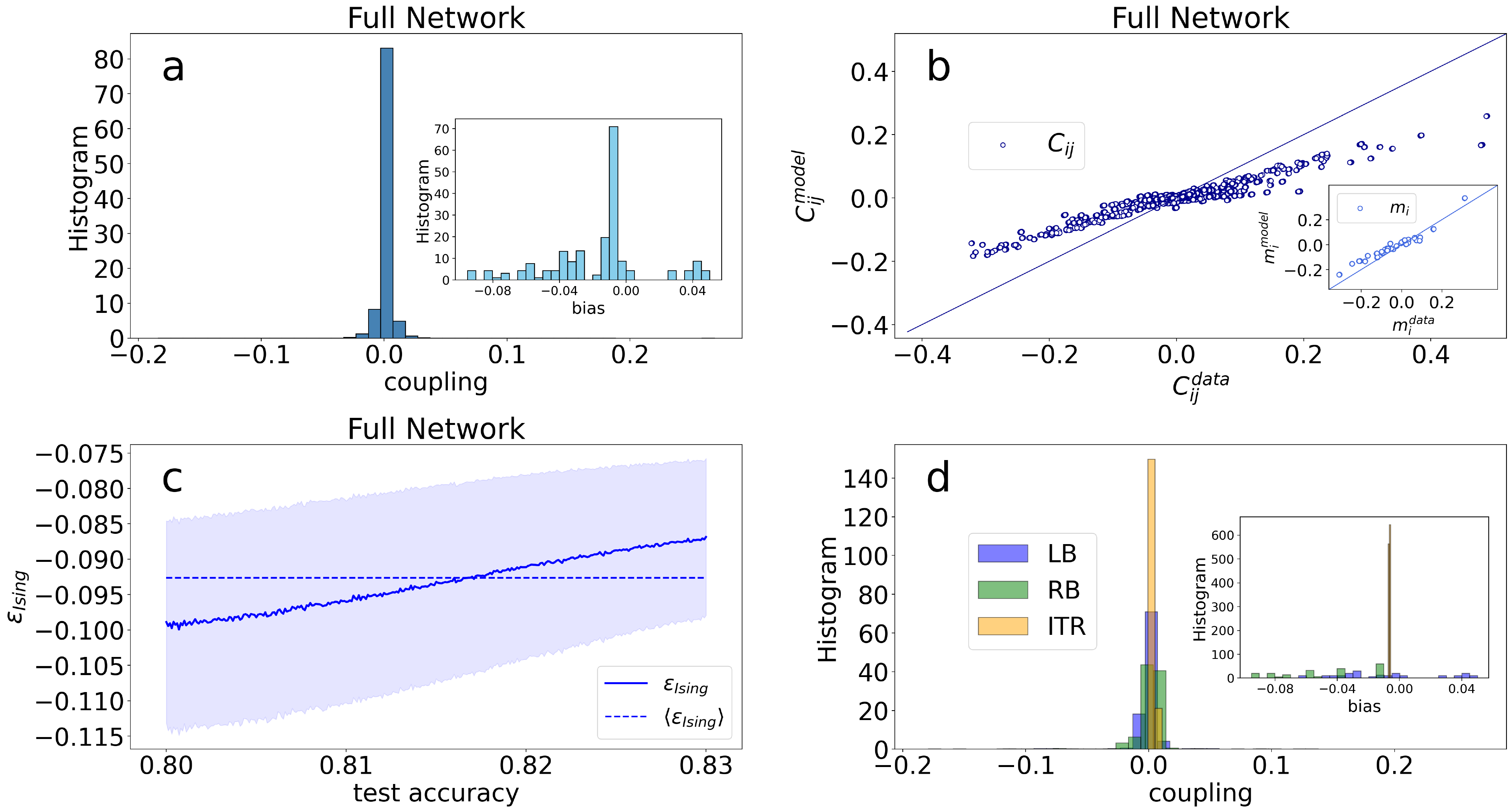}
	\caption{Properties of the data-driven Ising model ($P=60\ 000$, $N=675$). The Ising model is constructed by the maximum entropy method.
	The total area of the histogram is normalized to one.
(a) Histograms of full-model (all) parameters. Couplings $J_{ij}$ and biases $h_i$ (inset) are shown.
(b) Reconstructed correlation $C_{ij}$ and magnetization $m_i$ (inset) from the effective Ising model versus 
the measured ones for the full model. Full lines indicate the equality. 
(c) Ising energy density $\epsilon_{\rm Ising}$ versus test accuracy of the full model. 
Each accuracy corresponds to a large number of weight configurations, 
where the blue solid line indicates expectation values and the light blue shadow denotes the fluctuation. 
The blue dashed line indicates the typical energy of the Ising model.
(d) Histograms of model parameters in sub-parts (see Fig.~\ref{DNN}) of the deep hierarchy. Couplings $J_{ij}$ and biases $h_i$ (inset) are shown.}
\label{hist}
\end{figure*}

In Fig.~\ref{hist}(b), we verify that the reconstructed magnetizations and correlations 
from the data-driven Ising model are in good agreement with the measured ones, which ensures that the data-driven
Ising model is an effective model whose model parameters $\boldsymbol{\Omega}=\{\mathbf{J},\mathbf{h}\}$ capture statistical properties of the deep-learning weight space.
Moreover, we also find that the data-driven Ising model can reproduce a significant fraction of three-weight correlations (Appendix~\ref{three}).
The histogram of the model parameters 
is shown in Fig.~\ref{hist}(a). Note that most of the couplings are concentrated around zero value for the full model, despite a relatively less frequent tail. 
In contrast, the bias is broadly distributed. To get more insights about the model parameters, 
we plot the histogram for different sub-parts of the deep network [Fig.~\ref{hist}(d)]. Compared to the boundaries, i.e.,
LB and RB, the interior's 
couplings are more concentrated around zero value, and moreover the corresponding biases are all concentrated around $-0.0065$ 
corresponding to the highest peak in the histogram of Fig.~\ref{hist} (a), which clearly implies that weights 
in the interior of the deep network have weak correlations, and moreover the magnetization gets close to zero,
behaving like in a high temperature phase, as we shall prove semi-rigorously using the concept of entropy landscape.
Interestingly, the number of peaks in the LB's (resp. RB's) bias distribution approximately
corresponds to the number of input (resp. output) neurons. This observation demonstrates that the LB's and RB's
weights perform redundant encoding and robust decoding, respectively.

Given the model 
parameters of the constructed Ising model, we can compute the model energy of every weight configuration directly.
Fig. \ref{hist}(c) shows the model energy density versus the test accuracy. Though the fluctuation is large, 
we find that the energies of sampled weight-configurations distribute around the typical energy of the Ising model
and they have high accuracies, which helps to build an intuitive 
relationship between the model energy landscape and the deep-learning landscape.

To cross-check the qualitative properties of the constructed Ising model, we also apply the pseudolikelihood maximization 
method to construct the Ising model, relying on the entire configuration data, rather than only the first two moments of the statistics.
This method allows us to control
the sparsity of the effective Ising model, i.e., the number of null couplings.
Introducing the null couplings in the model is physically intuitive, since weights in different layers may not strongly interact with each other.
Therefore, we construct sparse Ising models only for deep networks.
Technical details are given in Appendix~\ref{app-e}.
As shown in Fig.~\ref{plmres}, the aforementioned properties are conserved. In fact, the null couplings explain the weakly interacting weights across layers,
corresponding to negligible values of couplings in the dense counterpart. Thus it can be expected that the qualitative behavior does not change for both methods.

\begin{figure*}
	\centering
	\includegraphics[bb=5 14 1762 564,width=\textwidth]{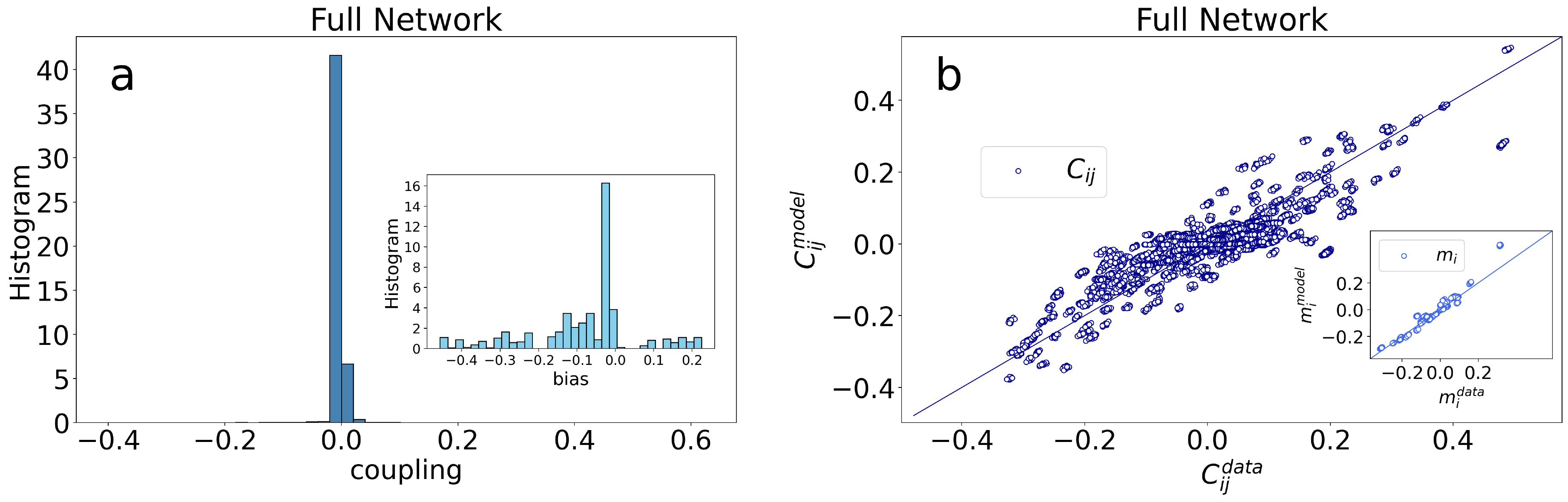}
	\caption{Properties of the data-driven Ising model ($P=60\ 000$, $N=675$). The Ising model is constructed by the pseudolikelihood maximization method ($\lambda=0.05$).
	The $\ell_1$ regularization training leads to a sparse model where about $73.5\%$ of weights are zero. 
(a) Histograms of full-model (all) parameters. Couplings $J_{ij}$ and biases $h_i$ (inset) are shown. The total area of the histogram is normalized to one.
(b) Reconstructed correlation $C_{ij}$ and magnetization $m_i$ (inset) from the effective Ising model versus 
the measured ones for the full model. Full lines indicate the equality. }
\label{plmres}
\end{figure*}

\section{Entropy landscape analysis}
\label{ELA}
In the previous section, we construct an effective Ising model, whose model parameters preserve 
the information up to second order correlations of the weight space. In order to explore the internal structure 
of the weight space of the deep learning, we introduce a distance-dependent 
term $x \sum_{i} \sigma_{i}^{*} \sigma_{i}$ in the original Boltzmann distribution [Eq.~(\ref{BZD})] as 
follows~\cite{Huang-2016}:
\begin{equation}
\label{Pentro}
P(\mathbf{\boldsymbol{\sigma} })=\frac{1}{Z}\exp \left( \beta\sum\limits_{i<j}{{{J}_{ij}}}{{\sigma }_{i}}{{\sigma }_{j}}+\beta\sum\limits_{i}{{h}_{i}}{{\sigma }_{i}}+x\sum\limits_{i}{\sigma _{i}^{*}}{{\sigma }_{i}} \right),
\end{equation}
where $\boldsymbol{\sigma}^*$ is the reference weight 
configuration taken from the collected weight data, its energy is computed from the Ising Hamiltonian, and $x$ is the coupling field constraining the distance 
between the configuration $\boldsymbol\sigma$ and the reference one $\boldsymbol\sigma^*$. 
Intuitively, $x>0$ implies that the weight configurations closer to the reference 
are preferred, while $x<0$ implies that the weight configuration far away from the reference are preferred (see Fig.~\ref{DNN} for an illustration). 

Then the geometric structure of the high-dimensional weight space can be summarized into the constrained free energy function given below. 
\begin{equation}
\beta f\equiv -\frac{1}{N}\ln \sum\limits_{\boldsymbol{\sigma} }{\exp }\left( -\beta E(\boldsymbol{\sigma} )+x\sum\limits_{i}{\sigma _{i}^{*}}{{\sigma }_{i}} \right),
\end{equation}
where $E(\boldsymbol{\sigma})$ is the Hamiltonian of the effective model.
We further introduce the energy density $\epsilon=E(\boldsymbol{\sigma})/N$ and the overlap $q=\sum_{i}\sigma_i\sigma_i^*/N$. It then follows that
\begin{widetext}
\begin{equation}\label{eq:f}
\begin{split}
\beta f & =-\frac{1}{N}\ln \iint{d}\epsilon dq\exp \left(-N\beta \epsilon +Nxq+\ln \sum\limits_{\boldsymbol{\sigma} }{\delta }\left(\epsilon -\frac{E(\boldsymbol{\sigma })}{N}\right)\delta \left(q-\frac{\sum\nolimits_{i}{{{\sigma }_{i}}\sigma _{i}^{*}}}{N}\right)\right)  \\
  & =-\frac{1}{N}\ln \iint{d}\epsilon dq\exp\left (-N\beta f(\epsilon ,q)\right),
\end{split}
\end{equation}
\end{widetext}
where $\beta f(\epsilon ,q)\equiv \beta\epsilon -xq- s(\epsilon ,q)$ is 
the constrained free energy as a function of energy density $\epsilon$ and overlap $q$. This constrained free energy captures
the competition between the model energy and entropy. The entropy is defined by $s(\epsilon ,q)=1/N\ln \sum\limits_{\boldsymbol\sigma }{\delta }(\epsilon -\frac{E(\mathbf{\sigma })}{N})\delta (q-\frac{\sum\nolimits_{i}{{{\sigma }_{i}}\sigma _{i}^{*}}}{N})$,
measuring the log-number of the weight configurations with energy $N\epsilon$ and
overlap $q$ given a reference.

When the system size $N$ ($675$ in this paper) is large in Eq.~(\ref{eq:f}), the 
Laplace method can be used to approximate the integral by taking only the dominant contribution. Hence, $f \approx \min _{\epsilon, q} f(\epsilon, q)$. 
To estimate the minimal free energy density $f(\epsilon,q)$ and the corresponding saddle-point values $(\epsilon,q)$, 
we iteratively solve the mean-field equations based on the cavity approximation (Appendix~\ref{app-b}). Then, the entropy 
density $s(\epsilon,q)$ is calculated via the double Legendre transformation $s(\epsilon ,q)= -\beta f + \beta\epsilon -xq$. 
For the convenience in the further analyses, we introduce the Hamming 
distance $\left(N-\sum_{i} \sigma_{i} \sigma_{i}^{*}\right) / 2$, which counts the number of weights 
that are distinct between the target configuration $\boldsymbol{\sigma}$ and the reference 
one $\boldsymbol{\sigma^*}$. The Hamming distance per weight $d$ can thus be transformed from 
the overlap by $d=(1-q)/2$. In the following analysis, we omit the energy density dependence 
of $s(\epsilon,q)$ and write it as the function of $d$, i.e., $s(d)$. We also assume $\beta=1$, as the inverse temperature has been
absorbed into the inferred couplings and biases of the effective model (see Sec.~\ref{Ising}).

In the case of sparse Ising models learned by the pseudolikelihood maximization method, the cavity approximation may break.
However, to estimate the entropy, one can use the heat-capacity integration technique (see Appendix~\ref{app-h}), despite its demanding computational cost.

To have a complete understanding of the deep-learning landscape, we analyze three representative scenarios.
The first one is the under-parametrization case, in which the number of training examples 
is larger than the network-parameter size $N$. The second one is the over-parametrization 
case where the number of training examples is smaller than the network-parameter size $N$, and the last one is the shallow-network 
case where all hidden layers are deleted (for more details, see the Appendix~\ref{app-c}).

We further define the ratio between the number of network parameters and the number of training examples as $\alpha=N/P$.
The point of $\alpha=1$ is a balance point where the number of network parameters matches exactly with the number of training examples.
In the deep learning of real-valued weights, with increasing values of $\alpha$, the test error profile displays the 
double-descent behavior~\cite{Belkin-2019,Neal-2018}, where an interpolation threshold marks the model complexity where the training 
error starts to vanish, and at the same location, a peak of test error emerges. The interpolation point $\alpha\sim\mathcal{O}(1)$, and the regime 
around the interpolation threshold is named the critical regime.
In contrast, we never observe the double descent behavior around $\alpha\sim\mathcal{O}(1)$ for 
deep learning of binary weights (see details in Appendix~\ref{app-f}).

\begin{figure*}
	\centering
	\includegraphics[bb=6 2 2333 548,width=\textwidth]{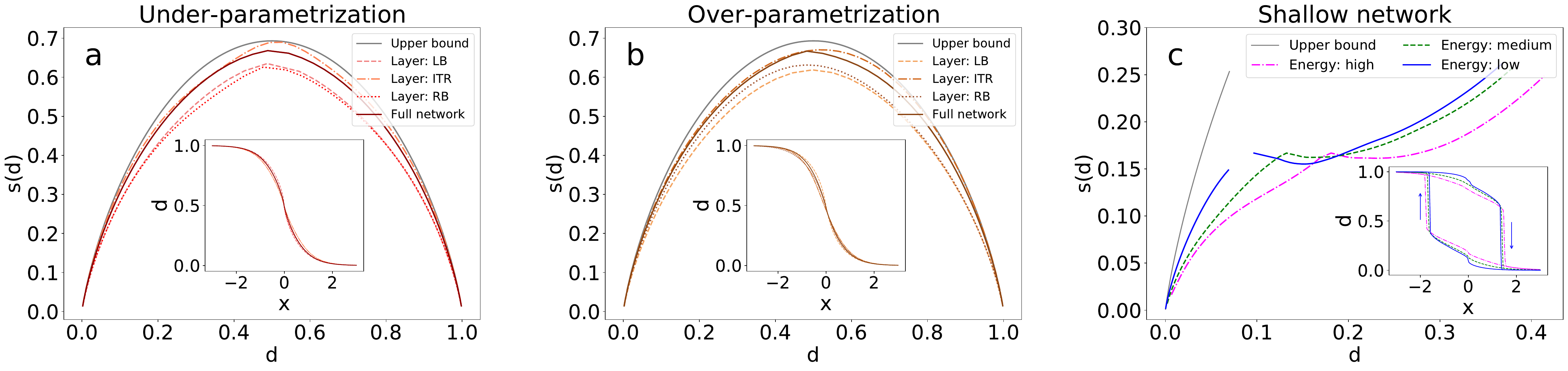}
	\caption{Entropy landscape analyses in three different scenarios of deep learning.
(a) Under-parametrization case ($\alpha\approx0.01$). The curves are the mean results 
over $20$ distinct reference weight configurations 
$\boldsymbol{\sigma}^*$ selected from the low energy region (for references coming from high- and medium-energy 
regions, see Fig.~\ref{S1} and Fig.~\ref{S2} in Appendix~\ref{app-c}). Entropy curves are plotted along the 
direction of the decreasing coupling field $x$ from $3$ to $-3$, while the Hamming distance 
per weight $d$ versus $x$ is plotted by considering two directions of increasing and decreasing $x$ (inset). 
No 
hysteresis loops and entropy gaps are observed. (b) Over-parametrization case ($\alpha\approx1.35$). Other conditions are the same as (a).
(c) Shallow network case. The inset shows a hysteresis loop for distance $d$ versus coupling field $x$, 
and the entropy curves correspond to the lower-branch of the hysteresis loop. A first-order phase 
transition arises in this case showing a complex landscape in the high-dimensional weight space.}
\label{entropy}
\end{figure*}

\begin{figure*}
	\centering
	\includegraphics[bb=4 3 1525 548,width=\textwidth]{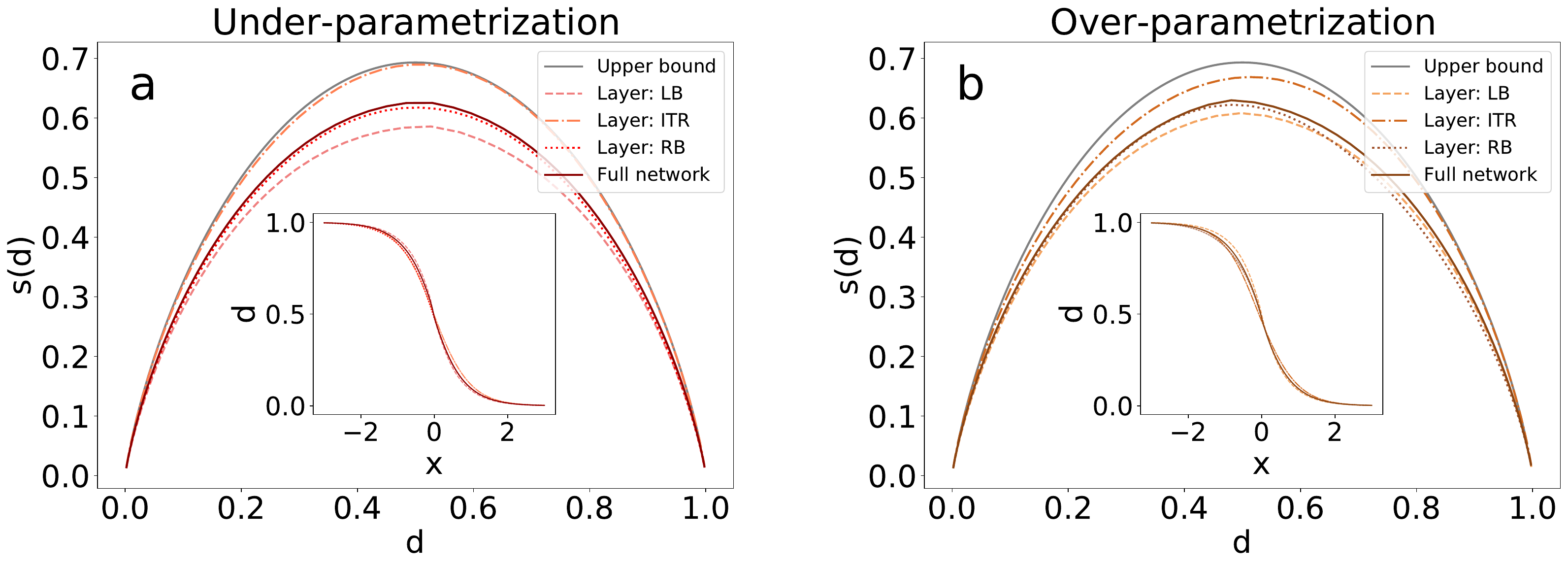}
	\caption{Entropy landscape analyses of the sparse Ising model (constructed by the pseudolikelihood maximization method, and $\lambda=0.05$).
	Other model settings are the same as Fig.~\ref{plmres}.
(a) Under-parametrization case ($\alpha\approx0.01$). The curves are the mean results 
over $20$ distinct reference weight configurations 
$\boldsymbol{\sigma}^*$ selected from the low energy region (for references coming from high- and medium-energy 
regions, the profile holds qualitatively). Entropy curves are plotted along the 
direction of the decreasing coupling field $x$ from $3$ to $-3$, while the Hamming distance 
per weight $d$ versus $x$ is plotted by considering two directions of increasing and decreasing $x$ (inset). 
No 
hysteresis loops and entropy gaps are observed. (b) Over-parametrization case ($\alpha\approx1.35$). Other conditions are the same as (a).}
\label{entropse}
\end{figure*}

\subsection{Under-parametrization scenario}
In the under-parametrization case, the number of supplied training examples is much larger than the network-parameter size (or the model complexity).
Interestingly, we do not find any weight-space-structure induced phase-transition.
First, the distance $d$ increases or decreases continuously following either of two 
directions of changing the coupling field [Fig.~\ref{entropy} (a)]; both directions collapse into a single curve, without any hysteresis phenomenon.
This provides a strong evidence that the weight space is connected in the absence of thermodynamically-dominant barriers. This result of the effective model coincides roughly with 
the low-dimensional projection evidence shown in Fig.~\ref{tsne} (b) as well.

Let us then look at the entropy landscape.
According to Eq.~(\ref{Pentro}), the entropy of the system at $x=0$ gives a rough estimate of the size of the original weight space, which is one advantage of 
our statistical mechanics analysis (see a similar study of constraint satisfaction problems~\cite{Huang-2012}).
By varying the value of $x$, we found that the entropy landscape is also smooth. For a system with discrete degrees of freedom, one can easily compute the upper bound for the entropy
$s_{ub}(d)=-d\ln d-(1-d)\ln(1-d)$ with the unique constraint of the distance between any two weight configurations in the high-dimensional space.
One salient feature is that three sub-parts of the deep architecture display distinct behaviors, despite a common smooth landscape.
Surprisingly, the interior region shows an entropy nearly saturating the upper bound. This region in physics behaves like a liquid state, where
there exist many realizations of weights to fulfill the classification task of the deep network, and the learning dynamics is thus fast in this region.
However, the right or left boundaries are more constrained than the interior region with lower entropy values. The left boundary is the most constrained
one among all three parts of the deep network, while the right boundary behaves similarly to the full network of three parts. An intuitive interpretation is that,
two boundaries are directly responsible for encoding or decoding the input-output relationship embedded in the training dataset. Allowing more freedoms in these two boundaries will
sacrifice the generalization capability of the network, which is also evidenced in our recent work of learning credit assignments~\cite{Li-2020}.
In sum, the under-parametrization does not lead to a broken weight space, with a liquid-like interior next to two 
more constrained boundaries.

The aforementioned properties of the deep learning weight space are qualitatively conserved, when the effective model is constructed by
applying the pseudolikelihood maximization method, where the sparsity of the model is controlled (Fig.~\ref{entropse}). When the cavity approximation breaks,
the heat-capacity integration method can be used, yielding the smooth landscape as well (data not shown).

The entropy landscape analysis also allows to compute the iso-energy curve (technical details are given in Appendix~\ref{app-b}).
Figure~\ref{iso} shows the iso-energy entropy curve, where every point on the curve shares the same energy with the reference configuration.
This curve is suppressed when $d\in[0.1,0.9]$, compared to the original entropy landscape. In the regime close to the reference, the entropy is nearly the same 
as that without the energy constraint. In the regime around $x=0$, there appears a negligible hysteresis loop, indicating that the roughness of the 
landscape is weak even if the iso-energy constraint is enforced.
Note that there exist two local minima in the energy-distance curve, whose asymmetry is caused by the external bias of weights.
This shows that by applying the current physics tool, one can access the structure of the specified
landscape around any reference point in the high-dimensional deep-learning space. 

\begin{figure}
\centering
\includegraphics[bb=3 3 1436 514,width=\textwidth]{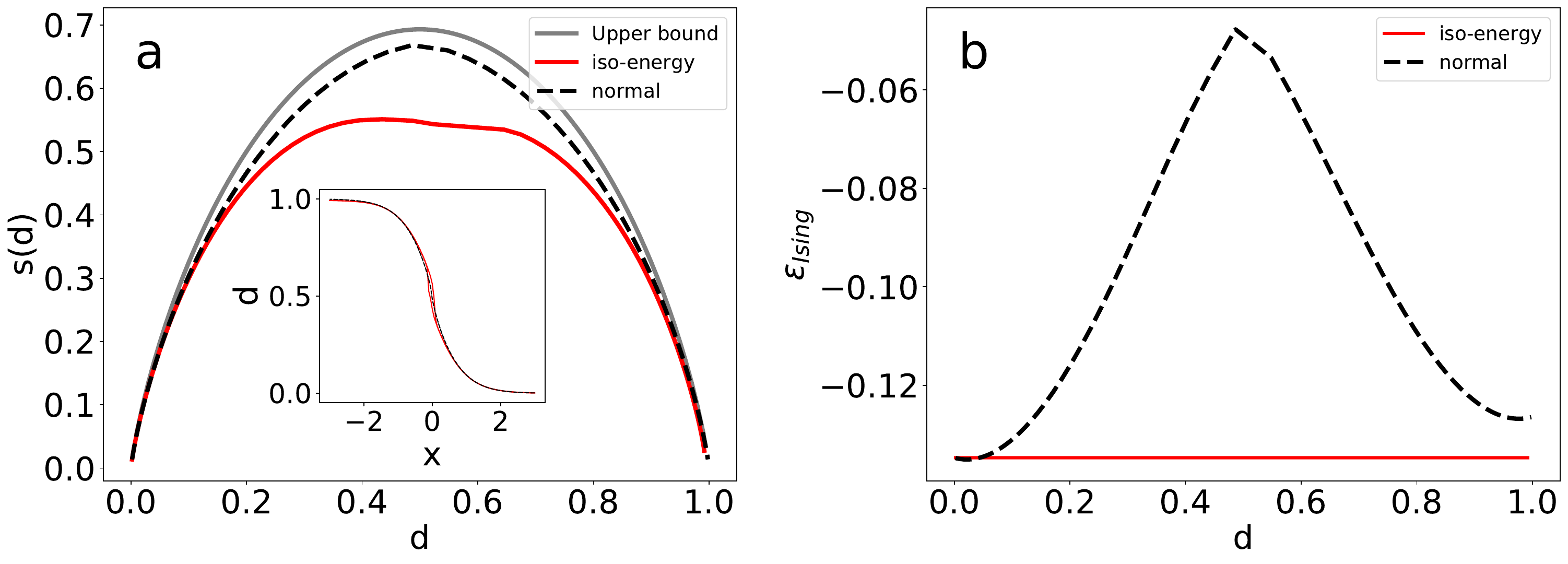}
\caption{Iso-energy entropy profile for the under-parameterization case. (a) 
The curves are mean results estimated from ten independent trials where the reference configurations $\boldsymbol{\sigma}^*$ are selected 
from the low energy region of $\epsilon_{\rm Ising}$. Both the iso-energy case and the normal 
case (no fixed-energy constraint) share the same reference configurations. (b) The iso-energy constraint
is achieved by fixing the energy to the Ising energy density $\epsilon_{\rm Ising}$ of 
the reference configuration, for which the secant method is applied to search for a compatible inverse temperature $\beta$, while
the coupling field $x$ 
still varies from $3$ to $-3$. In the normal case, the inverse temperature $\beta$ is set to $1$.
$d=0.5$ (or $x=0$) corresponds to the original unperturbed model.
} \label{iso}
\end{figure}

\subsection{Over-parametrization scenario}
\label{subs-op}
We then ask whether the nice property of the under-parametrization case transfers to the over-parametrization case, which is the popular setting in the current deep learning era.
We thus 
investigate the over-parametrization case, where the number of training examples is less than the model complexity (Appendix~\ref{app-c}).
The standard over-parameterization requires that $\alpha\gg1$, while our analysis is limited to a light version ($\alpha>1$) of over-parametrization.
We surprisingly find that the qualitative properties of the under-parametrization case also hold in the over-parametrization case [Fig.~\ref{entropy} (b)], in an
excellent agreement with recent empirical studies of deep learning at a large-scale architecture~\cite{Gari-2018,Draxler-2018,Nguyen-2019}.
All these previous studies confirmed that the stochastic gradient descent dynamics does not suffer from high loss
barriers~\cite{compdyn-2018}, and the loss function landscape is characterized by many minima yet separated by low barriers~\cite{Lee-2020}.
Our finding adds further evidences by claiming a smooth landscape for deep networks of binary synapses, in both under- and over-parametrization 
scenarios, which may be connected to the absence of double descent phenomena (Appendix~\ref{app-f}).

However, we notice
a salient difference that the extent the interior entropy gets close to the upper bound is weaker than that of the under-parametrization case, 
despite the fact that the interior entropy is still highest among three parts of the deep network. Even in the over-parametrization case,
a liquid-like interior inside the network exists, coinciding exactly with a recent study of deep networks in both random inputs/outputs setting and
teacher-student generalization setting~\cite{Yoshino-2020}. This recent study claimed a glass-liquid-glass/solid-liquid-solid pattern
inside the deep-learning architecture. Despite a relatively small scale for the sake of the data-driven analytical study, our work claims the same phenomenon in the central part of the deep 
network by constructing an effective data-driven model for a practical deep learning. Although we do not observe a glass/solid like boundaries (unlike in the recent study~\cite{Yoshino-2020}),
these two boundaries are clearly more constrained than the central part. We thus expect that these boundaries may enter a glass/solid phase in the thermodynamic limit, which is 
a very interesting future direction.

\subsection{Shallow networks}
To explore the benefit of depth, an important characteristic of the deep learning, we study a shallow-network architecture without any hidden layers.
Other conditions are the same with the other two scenarios. Note that, in terms of test performance, adding hidden layers of increasing width improves over the shallow counterpart (Appendix~\ref{app-g}).
Using the same statistical mechanics framework, we surprisingly find a first-order phase 
transition in the distance-coupling-field profile [Fig.~\ref{entropy} (c)]. The discontinuous transition is characterized by
a hysteresis loop and associated entropy gaps, irrespective of references selected from high-, medium- and low-energy regions.
For the shallow-network case,
the weight space is broken into clusters separated by entropy gaps. In particular, a double-discontinuous transition is even observed for the low-energy references, implying that
the weight space around the low-energy references becomes much more complex, as also revealed in a biological data analysis of retina coding~\cite{Huang-2016}.

Compared to the deep network, the dynamics in the shallow network can be strongly affected by the discontinuous transition because of metastability in thermodynamic states.
When adding more hidden layers, the broken weight space is replaced by a connected smooth sub-space, demonstrating the benefit of depth. 
Our effective model thus distinguishes a shallow network from its deep counterpart, as also supported by the Monte-Carlo dynamics in the next section.

\begin{figure}
\centering
\includegraphics[bb=4 1 1772 408, width=\textwidth]{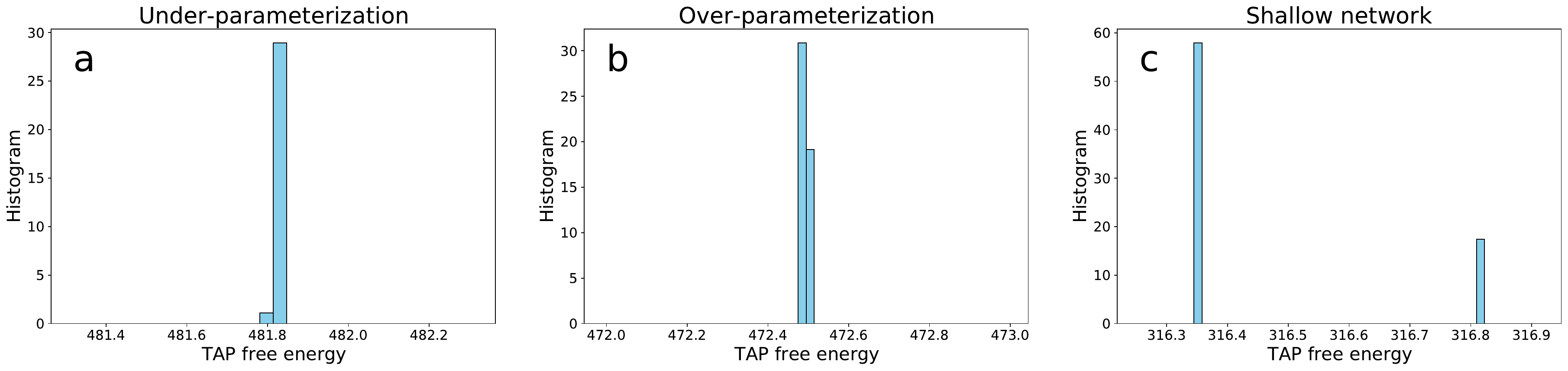}
\caption{Distribution of TAP free energies of the Ising model. The model is constructed by the maximum entropy method.
The histogram is obtained by iterating the TAP equation (see Appendix~\ref{app-i}) from $2\ 000$ random initializations of 
the magnetizations.
(a) Under-parametrization case. (b) Over-parametrization case. (c) Shallow network case.}
\label{deepeak}
\end{figure}

\begin{figure}
\centering
\includegraphics[bb=7 7 1729 528, width=\textwidth]{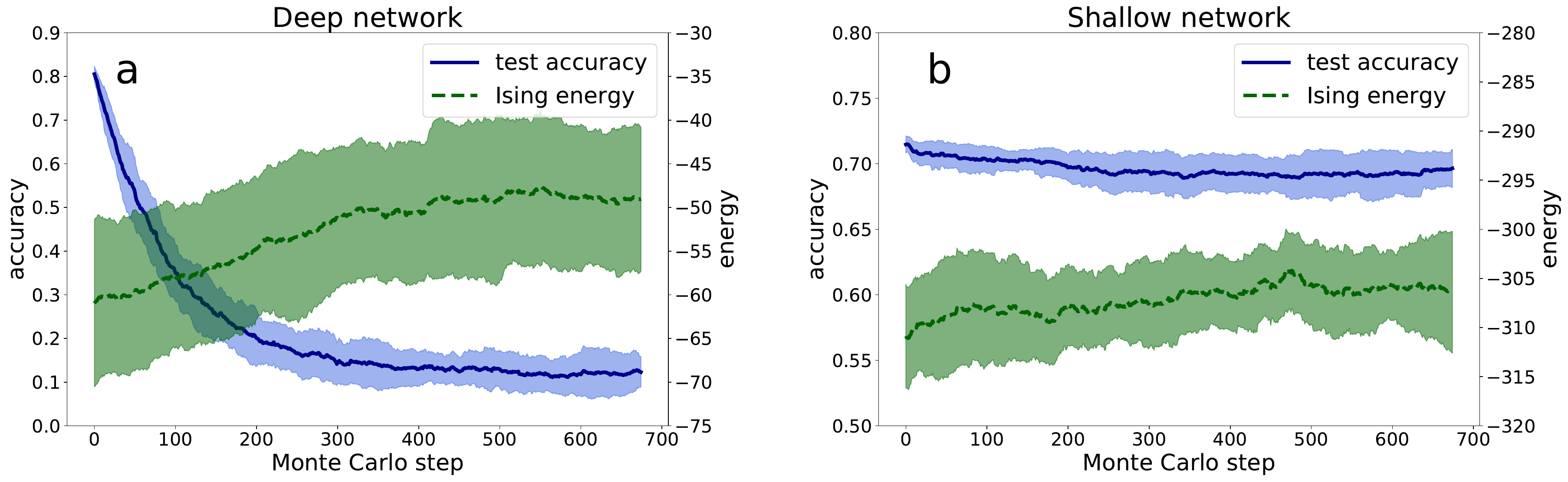}
\caption{Monte Carlo dynamics on the landscape of the Ising model. 
The dynamics starts at a reference weight configuration, and is measured in a time unit of one proposed spin flip (one step in the abscissa).
The test accuracy and Ising energy are both tracked during the dynamics. The fluctuation is computed from $100$ independent trials.
(a) Deep network case. (b) Shallow network case.}
\label{deepMC}
\end{figure}

\begin{figure*}
	\centering
	\includegraphics[bb=4 1 1436 983,width=\textwidth]{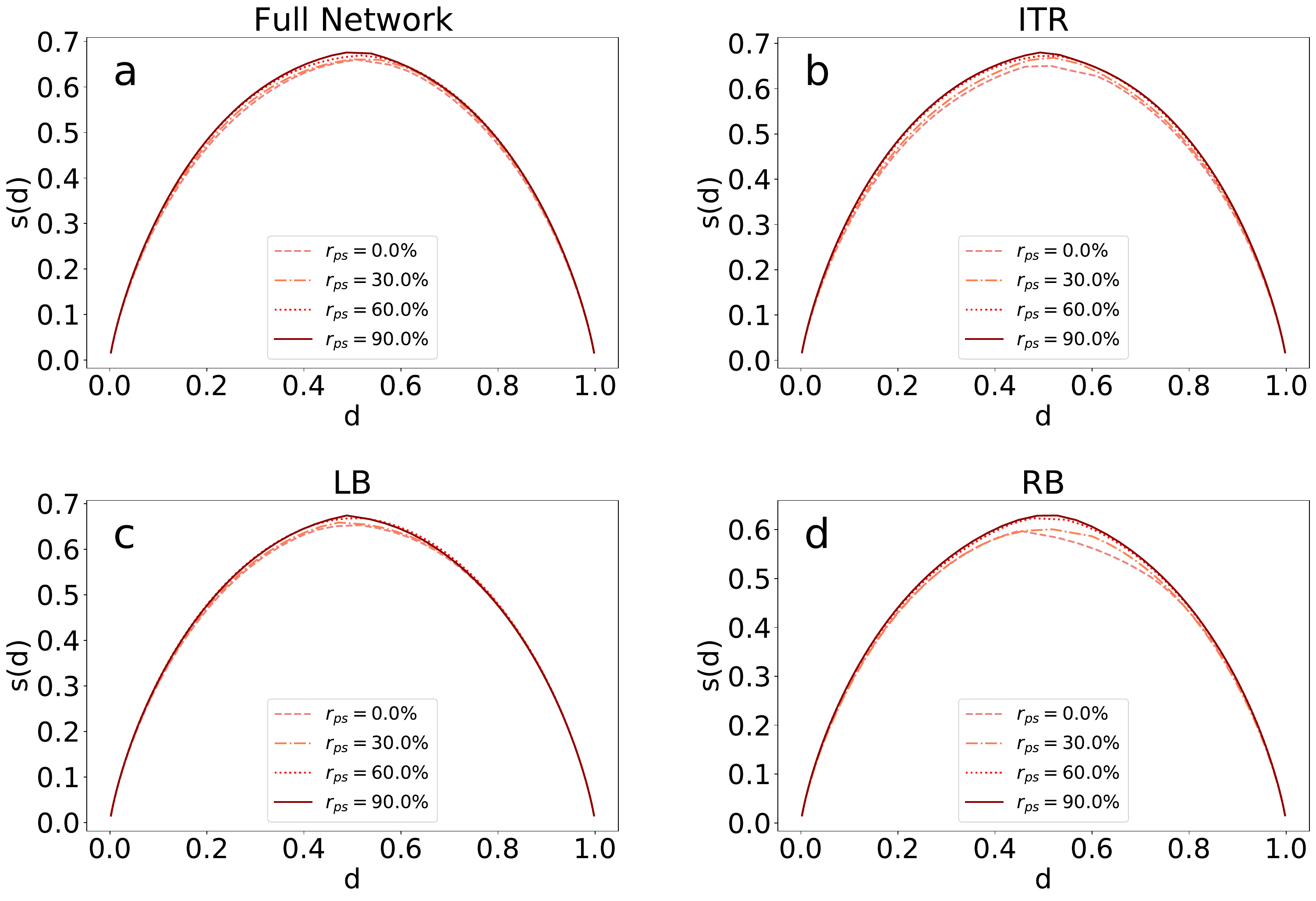}
	\caption{Effects of permutation symmetry on the entropy landscape ($P=60\ 000$, $N=675$). The Ising model is constructed by the maximum entropy method.
	The procedure of generating permutation symmetric weight configurations is detailed in Appendix~\ref{app-j}.
	The fraction of permutation symmetry ($r_{\rm ps}$) is also specified there, and indicates how large the number of the permutation-dependent weight configurations (used for learning
	the Ising model) is. The entropy is the mean value over $20$ independent references.
(a) Full network.
(b) Interior part of the network.
(c) Left boundary of the network.
(d) Right boundary of the network.}
\label{psres}
\end{figure*}

\subsection{Distributions of TAP free energy and Monte-Carlo dynamics of the effective model }
To verify the geometric picture gained from the entropy landscape, we study the TAP free energy distribution of the effective
Ising model (see Appendix~\ref{app-i} for technical details). As shown in Fig.~\ref{deepeak} for the under-parametrization setting, the distribution
shows a single peak, consistent with the smooth landscape. The over-parameterization setting has similar results.
The single peak profile is further corroborated by the Monte-Carlo dynamics (like that in Appendix~\ref{three}) starting from a reference solution state.
As shown in Fig.~\ref{deepMC}, the local dynamics is apt to drift towards chance-level configurations, without difficulty of
crossing high energy-barriers. In contrast, the Monte-Carlo dynamics of the Ising model for the shallow network gets trapped by the basin where the reference state
sits on, which coincides exactly with the first-order transition (i.e., emergence of high energy barriers in the phase space).
This observation is also supported by the double-peak profile of TAP free energy distributions (Fig.~\ref{deepeak}).
The above properties are also observed in the sparse Ising model of deep learning.

To sample one high-accuracy weight-configuration, one can instead use Eq.~(\ref{eq:p}), as the probability parameters are directly trained, while the effective model captures only up to
pairwise correlations of the weight-space statistics, showing a smooth landscape for deep networks, and thus the local dynamics is easy to escape from any solution state.
The effective model could predict a significant fraction of three-weight correlations, and the physics interpretation leads to consistent pictures about the weight space as well as
the learning, and even inspire algorithmic designs (shown below). Therefore, the constructed Ising model could be an effective model. It would be very interesting to explore in future works whether 
higher-than-second order correlations in the data enhance our current understanding.

\subsection{Effects of node-permutation symmetry on the deep learning landscape}
There exists node-permutation symmetry in deep learning, which is able to relate one weight configuration to another one 
via a permutation of hidden nodes (see Appendix~\ref{app-j}). To explore effects of node-permutation symmetry, we design a procedure of ranking 
the pre-activations of hidden layers, and then control the fraction of permutation symmetry in the collected weight configurations.
As shown in Fig.~\ref{psres}, increasing the permutation symmetry fraction leads to an enlarged entropy profile, for four different network settings.
Two key points must be remarked. First, adding the permutation symmetric configurations for learning the effective model does not affect the qualitative behavior of the 
entropy profile. Second, the close-by regime of any reference is not affected by the fraction of the permutation symmetric configurations. According to this observation, we conjecture that
the underlying picture is that the deep learning weight space under investigation is smooth as a whole.

\subsection{Algorithmic implication of the effective model}
\begin{figure*}
	\centering
	\includegraphics[bb=3 4 743 401,width=\textwidth]{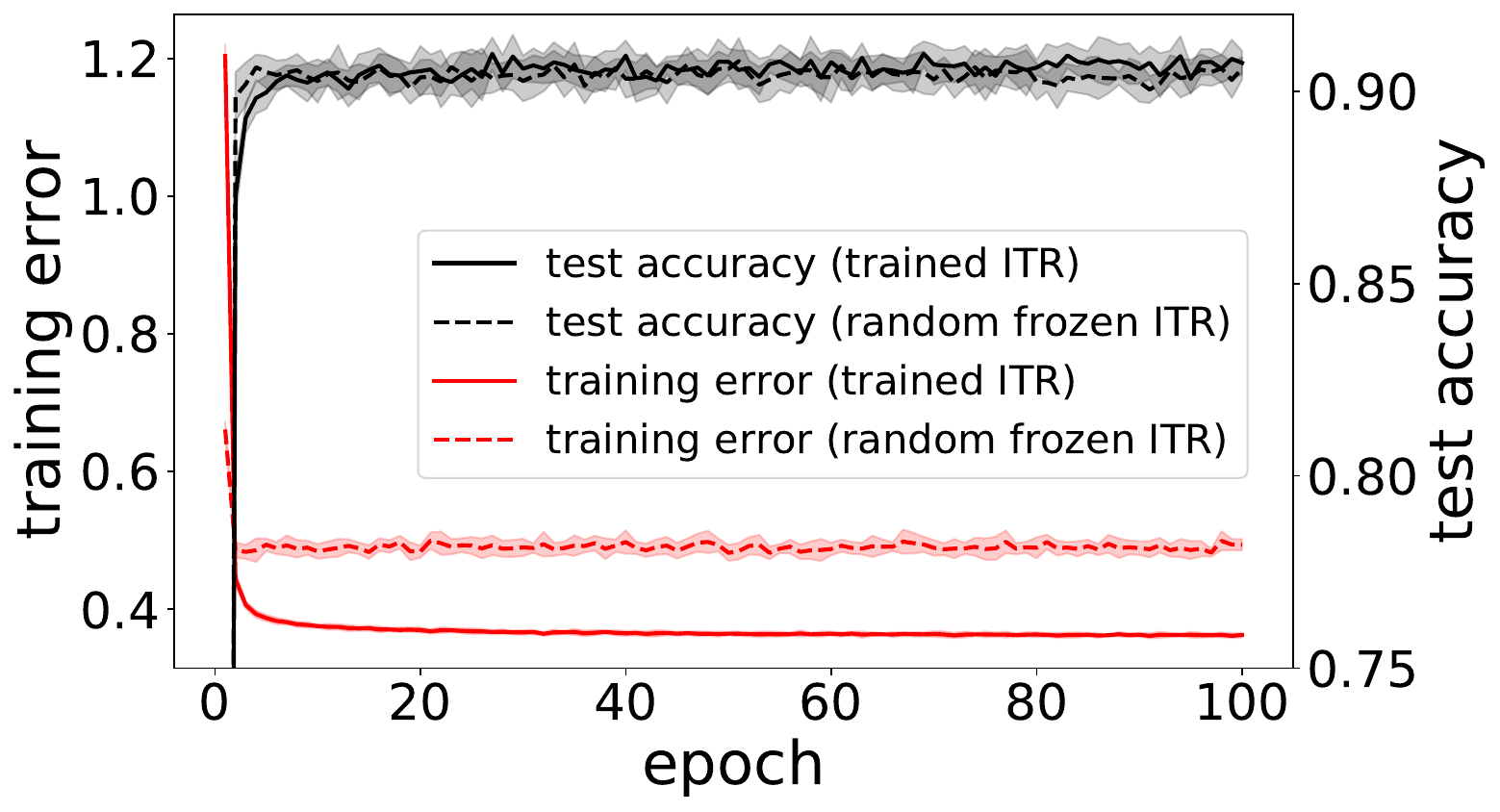}
	\caption{Algorithmic implication of the effective model. Training with randomly-frozen ITR works, consistent with our theory. The training is carried out on the
	network architecture 784-100-100-10 with the full dataset. The performance of training with trained ITR is compared.}
\label{frozen}
\end{figure*}

Our theory predicts a liquid core inside the deep network. We ask a last question---does the effective model lead to an algorithmic design, despite the effective model built for 
small-scale networks? To address this question, we first randomly freeze the connections inside the ITR, i.e., each weight value is chosen with equal probabilities, and then its 
external field $\theta$ is quenched to a positive large value (if $w=+1$) or a negative large value (if $w=-1$). The other parts of the network are normally trained, as we did before.
We surprisingly find that this scheme works in practice (Fig.~\ref{frozen}), coinciding exactly with the liquid-like core hypothesis drawn from the effective model. An additional salient advantage is that
the training would 
become fast, in that the ITR does not need learning.
To conclude, although our effective model is data-driven by the weight interactions of small-scale deep networks, the derived insights could inspire algorithmic designs for large-scale networks.

\section{Conclusion}
Key properties of deep networks, including the learning dynamics and the generalization capability, 
can be affected by the geometry of the weight-space landscape. Here, we design a data-driven effective model, 
considering all three necessary practical issues of a deep learning---structured
data, deep architectures, and training, to explore the geometric structure of the entire weight space of
a deep-learning landscape. Our study provides an interpretable effective model of deep learning weight space
in small-size neural networks, although a theory for finite-size deep networks is rare in previous studies. Remarkably, the theory implies algorithmic designs for large-scale deep networks with fast training.

Our effective model reveals several interesting phenomena. First, the model can distinguish a deep network from a shallow one in terms of weight-space geometric structure.
The deep network weight-space is smooth and dominated by a connected component, while a broken weight space emerges in the shallow network, displaying a discontinuous transition in the Hamming distance profile.
Second, for the deep network, the under-parametrization and over-parametrization belong to the same class, in the sense that their weight spaces are identically dominated by a single connected component, which is consistent with
recent empirical studies that claimed no substantial barriers between minima in the deep-learning loss landscape~\cite{Gari-2018,Draxler-2018,Nguyen-2019}. 
Our statistical mechanics analysis thus provides insights towards understanding the success of deep learning despite the NP-hardness (worst case) of the deep learning problem. 

A recent work revealed that as the model complexity of continuous weights increases, 
the generalization error displays a cusp around a critical value of the complexity; on both sides of this cusp, the generalization error decays~\cite{Spigler-2019}. It is thus interesting in future works to see whether there exists a qualitative change in the geometric structure of the 
weight space around the cusp. We further remark that in our current study, the model complexity is fixed for simplicity. In fact, the model complexity can also be tuned to study the existence of the double descent phenomenon (see Appendix~\ref{app-f}).
We confirm that in the explored range of model complexity, the double descent phenomenon is absent for deep learning of binary synapses. In other words,
the generalization capability of the network continuously improves with increasing model complexity.

Third, a special interior part of a largest entropy in the weight space exists, showing a liquid-like property.
The liquid phase is defined with a vanishing glass-order-parameter~\cite{Yoshino-2020}, or the absence of ergodicity breaking.
In this sense, the learning dynamics in the interior part must be fast, because both directions of each weight are not highly biased, or close to
a random initialization.
In the particular scenario of under-parametrization, 
this giant interior entropy even gets very close to the upper bound of entropy set by a completely random-weight space.

The existence of the liquid-like central part coincides exactly with two recent studies, despite the current focus of binary weights.
One is the toy random models of deep learning~\cite{Yoshino-2020}, as discussed in Sec.~\ref{subs-op};
the other is the 
ensemble backpropagation applied to practical deep learning~\cite{Li-2020}, which found a peak of model entropy in the central part of the deep network. 
We remark that these two works studied continuous degrees of freedom and 
are restricted to only the over-parametrization scenario, while our current effective model focuses on discrete degrees of freedom, reducing a hard-to-describe practical deep network to
a theoretically-amenable Ising model, which establishes a bridge between abstract physics model and practical deep learning.
Based on our model, the global geometric structure of any specific point in the high-dimensional weight space can be explored, while this is hard 
for those toy models or numerical experiments.

Our work also provides several significant insights about our current understanding of deep networks with discrete synapses. First, the role of depth
can be interpreted in characteristics of the high-dimensional weight space. Second, the under-parameterization and the over-parameterization
make no difference in the qualitative property of the weight space. Third, deep learning shapes a liquid-like core in the weight space for the central part of the architecture, 
which suggests that there exists a fast mode for the learning dynamics along the hierarchy, thereby inspiring further algorithmic development. 
Although it is still challenging to derive an effective model of large-scale deep learning, we anticipate that the insights gained from this study,
including smooth landscapes, liquid-like core, TAP free energies, effects of permutation symmetry, and absence of double descent phenomenon, would inspire derivation of general principles governing 
the relationship among learning dynamics, geometry of weight space, and the behavior of generalization. In particular, it remains open what the landscape looks like around the test error cusp for 
modern deep learning of real-valued weights, which our framework may help to clarify, and we will explore this interesting direction in future works.

\begin{acknowledgments}
We would like to thank other members of PMI lab for a long-term discussion of this project. This research was supported by
the National Key R$\&$D Program of China
(2019YFA0706302), the National Natural Science Foundation of China for
Grant No. 11805284, and the start-up budget 74130-18831109 of the 100-talent-
program of Sun Yat-sen University.  
\end{acknowledgments}
\onecolumngrid
\appendix
\section{Mean-field training algorithms}
\label{app-a}
To train a deep supervised network with binary weights, we apply the mean-field method first introduced in the previous work~\cite{Shayer-2018}. 
In our current setting, each weight $w^l_{ij}$ is sampled from 
a binomial distribution $P(w^l_{ij})$ parametrized by an 
external field $\theta^l_{ij}$ as follows,
\begin{equation}\label{eq:p2}
P(w^l_{ij}) = \sigma(\theta^l_{ij})\delta_{w^l_{ij},1} + [1-\sigma(\theta^l_{ij})]\delta_{w^l_{ij},-1},
\end{equation}
with mean $\mu^l_{ij}=2\sigma(\theta^l_{ij})-1$ and variance 
$(\sigma^l_{ij})^2=-4\sigma^2(\theta^l_{ij})+4\sigma(\theta^l_{ij})$. 
According to the central-limit theorem, the feedforward transformation 
can be re-parametrized as:
\begin{subequations}
\begin{align}
&z_j^l = m_j^l + v_j^l\cdot\epsilon_j^l ,\\
&a_j^l = \frac{1}{\sqrt{N_{l-1}}} \operatorname{L-ReLU}(z_j^l),
\end{align}
\end{subequations}
where $m_j^l=\sum_i\mu^l_{ij}a_i^{l-1}$, and $v_j^l=\sqrt{\sum_i(\sigma^l_{ij})^2 (a^{l-1}_{i})^2}$.

During the error backpropagation phase, we need to compute the gradient of the loss 
function $\mathcal L$ [e.g., $L_{{\rm CE}}$ in Fig.~\ref{tsne} (a)] with respect to the external field $\boldsymbol{\theta}$, which proceeds as follows:
\begin{equation}
\frac{\partial \mathcal L}{\partial \theta_{ij}^l}=\frac{\partial \mathcal L}{\partial z^l_j}\frac{\partial z^l_j}{\partial \theta^l_{ij}}=\frac{\partial \mathcal L}{\partial z^l_j}\Bigg(\frac{\partial m^l_j}{\partial \theta^l_{ij}}+ \epsilon^l_j \frac{\partial v^l_j}{\partial \theta^l_{ij}}\Bigg).
\end{equation}
We then define $\Delta_j^l \equiv \frac{\partial \mathcal L}{\partial z^l_j}$. 
On the top layer, $\Delta_j^l = y_j^L - \hat y_j^L$, 
where $y_j^L = \frac{e^{z^L_j}}{\sum_i e^{z^L_i}}$ is the softmax output, and $\hat y_j^L$ is the (one-hot) label of the input. 
On the lower layers, given $\Delta_k^{l+1}$, we can iteratively compute $\Delta_j^l$:
\begin{equation}
\begin{split}
\frac{\partial \mathcal L}{\partial z^l_j} 
&= \sum_k \frac{\partial \mathcal L}{\partial z^{l+1}_k}\frac{\partial z^{l+1}_k}{\partial z^l_j}\\
&= \sum_k \Delta_k^{l+1}f'(z^{l}_j)\Bigg(\mu^{l+1}_{jk}+\epsilon^{l+1}_k\frac{(\sigma^{l+1}_{jk})^2 a^l_j }{v^{l+1}_k}\Bigg).
\end{split}
\end{equation}
Finally, we compute $\frac{\partial m^l_j}{\partial \theta^l_{ij}}$ 
and $\frac{\partial v^l_j}{\partial \theta^l_{ij}}$, respectively. It then proceeds as follows.
\begin{subequations}
\begin{align}
&\frac{\partial m^l_j}{\partial \theta^l_{ij}}=\frac{\partial m^l_j}{\partial \mu^l_{ij}}\frac{\partial \mu^l_{ij}}{\partial \theta^l_{ij}}= 2 a^{l-1}_i \sigma'(\theta^{l}_{ij}),\\
&\frac{\partial v^l_j}{\partial \theta^l_{ij}}=\frac{\partial v^l_j}{\partial (\sigma^l_{ij})^2}\frac{\partial (\sigma^l_{ij})^2}{\partial \theta^l_{ij}}= -2 \frac{(a^{l-1}_i)^2\mu^l_{ij}\sigma'(\theta^l_{ij})}{v^l_j}.
\end{align}
\end{subequations}
Note that $\boldsymbol{\epsilon}$ is sampled and quenched for both forward and backward computations in a single mini-batch gradient descent. During training, we add an $L_2$ regularization to controlling the magnitude of 
external fields.
After the learning is terminated, an effective network with binary weights can be constructed by sampling the binomial distribution parametrized by external fields.

\section{The Ising model predicts three-weight correlations}
\label{three}
\begin{figure}
\centering
\includegraphics[bb=2 3 1451 566, width=\textwidth]{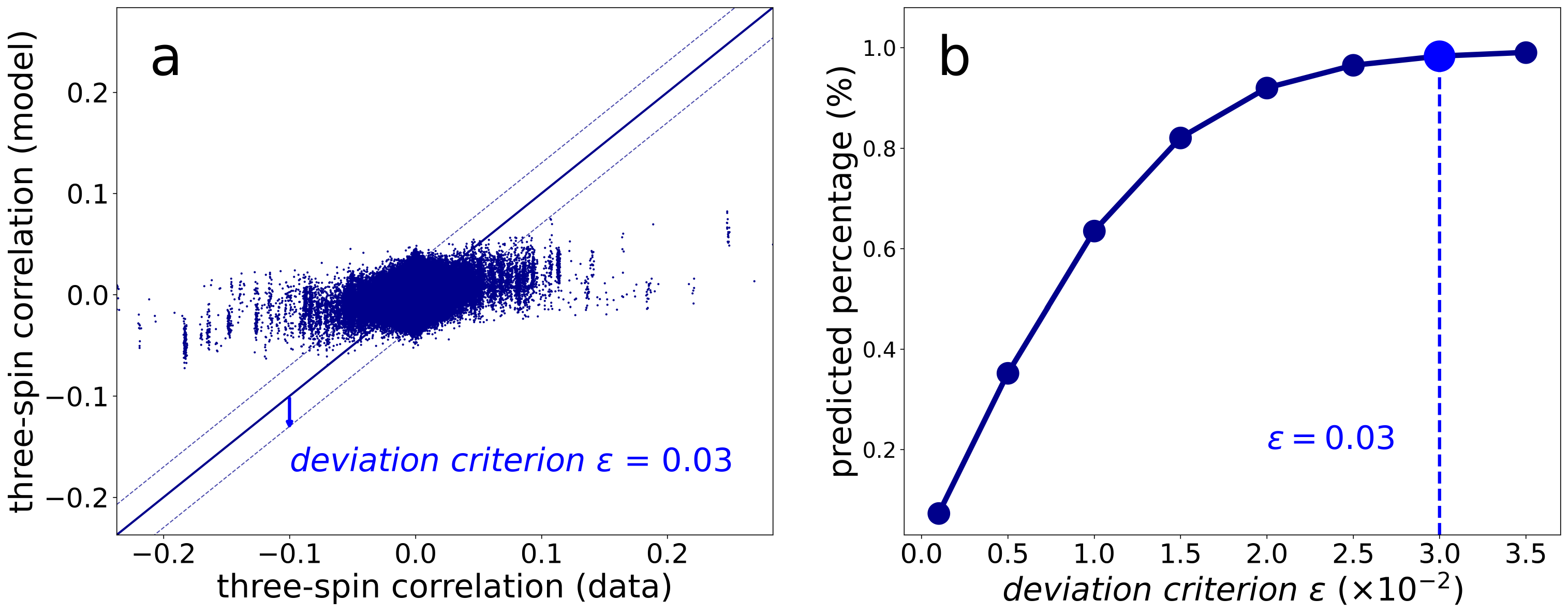}
\caption{Three-spin correlations of the effective Ising model, in comparison with those estimated from the collected weight ensemble.
(a) Predicted three-spin correlations versus the measured ones (data). $200$ spins are randomly 
selected from the effective model to do the comparison, due to a huge number of three-spin correlations. 
The full line shows equality, and dashed lines correspond to the 
deviation criterion $\epsilon$, which is the absolute value of the deviation between the predicted three-spin correlation and the true one.
The predicted three-spin correlation is estimated by running a Monte-Carlo simulation of the effective Ising model. 
(b) Predicted percentage versus deviation criterion. The predicted percentage in all considered three-spin correlations
is the percentage of those three-spin 
correlations whose absolute deviation from the true one is within the given deviation 
criterion $\epsilon$.	
}\label{MC-comp}
\end{figure}

To see whether the data-driven Ising model can reproduce the three-weight correlations, we first run a Monte-Carlo experiment on the Ising Hamiltonian,
and then carry out a sampling of the steady state space of the Monte-Carlo dynamics after a sufficient relaxation. The three-weight correlations are estimated from 
these samples, and are then compared with those estimated directly from the data (weight ensemble). The result is shown in Fig.~\ref{MC-comp}, which suggests that 
the Ising model can predict a significant fraction of three-weight correlations (e.g., about $82\%$ provided that the criterion is set to $\varepsilon=0.015$).

\section{Cavity method for landscape analysis}
\label{app-b}
To analyze the statistical properties of the least structured model, we apply the cavity method in the spin glass
theory~\cite{Mezard-2001}. Note that the weight configuration follows the Boltzmann 
distribution $P(\boldsymbol\sigma) = \exp\left(-\beta E(\boldsymbol\sigma)\right)/Z $, 
where the energy $E(\boldsymbol\sigma)=-\sum_{i<j} J_{i j} \sigma_{i} \sigma_{j}-\sum_{i} h_{i} \sigma_{i}$, 
and $Z$ is the partition function. An exact computation of the free energy ($-T\ln Z$) is an NP-hard problem.
To get an approximate free energy, we first compute the free energy change $\Delta F_i$ when 
adding a weight (e.g., $\sigma_i$) and its associated interactions. We then calculate the 
free energy change $\Delta F_a$ when adding an interaction (e.g., $J_{a}\sigma_i\sigma_j$, and $a\equiv(ij)$). In all these derivations, 
we make an assumption that when an interaction is removed, 
its
neighboring weights become independent such that the joint distribution of them becomes 
factorized, which is exact if the factor graph of the model is locally tree-like. The factor graph is a bipartite graph where two kinds of nodes (weight nodes and interaction nodes)
are present. For detailed derivations, interested readers are recommended to go through the standard textbook~\cite{MM-2009}, or the appendix of a recent paper~\cite{Huang-2017}.
In the cavity approximation, the free energy can be constructed as below,

\begin{subequations}
\begin{align}
 F &\equiv-\frac{1}{\beta}\ln Z =\sum_{i} \Delta F_i - \sum_{a}(|\partial a|-1) \Delta F_a,\\
\beta\Delta F_i &=-\ln \sum_{x=\pm 1} \mathcal{H}_{i}(x),\\
\mathcal{H}_{i}(x) &\equiv e^{x \beta h_{i}} \prod_{b \in \partial i} \cosh \beta J_{b}\left(1+x \hat{m}_{b \rightarrow i}\right),\\
\beta\Delta F_a &=-\ln \cosh\beta J_{a}-\ln \left(1+\tanh \beta J_{a} \prod_{i \in \partial a} m_{i \rightarrow a}\right),
\end{align}
\end{subequations}
where $|\partial a|$ denotes the number of weights connected to the
interaction $a$, $i\in \partial a$ denotes the neighboring weights $i$ of the
interaction $a$, and $a\in \partial i$ denotes the neighboring interactions $a$ of the
weight $i$. $m_{j\rightarrow b}$ is the cavity magnetization of weight $j$ in the 
absence of the interaction $a$, and $\hat{m}_{b \rightarrow i}=\tanh \beta J_{b} \prod_{j \in \partial b\backslash i} m_{j \rightarrow b}$ is 
the conjugate magnetization. 

According to the variational principle, the free energy should be stationary
with respect to $\{m_{i\rightarrow a}\}$. We can then obtain a set of self-consistent 
equations called message passing equations as follows,
\begin{subequations}
\begin{align}
m_{i \rightarrow a}&=\tanh \left(\beta h_{i}+\sum_{b \in \partial i\backslash a} \tanh ^{-1} \hat{m}_{b \rightarrow i}\right),\\
\hat{m}_{b \rightarrow i}&=\tanh \beta J_{b} \prod_{j \in \partial b \backslash i} m_{j \rightarrow b},
\end{align}
\end{subequations}
where $\backslash i$ means that the weight $i$ should be excluded from the set. $m_{i \rightarrow a}$ is interpreted as the message passing from weight $i$ to 
interaction $b$, and $\hat{m}_{b \rightarrow i}$ is interpreted as the message passing from 
interaction $b$ to weight $i$. By iterating these equations from some random initialization, the iteration will 
converge to a fixed point $\{m_{i \rightarrow a}^*\}$, which can be used to compute thermodynamic moments for learning (introduced below)
and entropy (used for landscape analysis). We remark that the fixed point corresponds to a local minimum of the free energy function. The algorithm may not converge 
when the model becomes glassy. 

We can directly compute the magnetization $m_i = \langle\sigma_i\rangle$, and 
correlation $C_{a}=\langle\sigma_i\sigma_j\rangle$ from the fixed 
point $\{m_{i \rightarrow a}^*\}$ as follows,
\begin{subequations}\label{Ising estimation}
\begin{align}
m_{i}&=\tanh \left(\beta h_{i}+\sum_{b \in \partial i} \tanh ^{-1} \left[\tanh \beta J_{b} \prod_{j \in \partial b\backslash i} m_{j \rightarrow b}^*\right]\right), \\
C_{a}&=\frac{\tanh \beta J_{a}+\prod_{i \in \partial a} m_{i \rightarrow a}^*}{1+\tanh \beta J_{a} \prod_{i \in \partial a} m_{i \rightarrow a}^*}.
\end{align}
\end{subequations}
These moments of the effective model can be used for updating the parameters of the model, approximating the model expectation
terms in the Boltzmann learning rule [Eq.~(\ref{eq:learning})], which are faster in our current setting than other methods, e.g., Monte-Carlo samplings.

Given the free energy, we can estimate the energy from the standard thermodynamics 
relation, $E = \frac{\partial(\beta F)}{\partial\beta}$, which is given by:
\begin{subequations}
\begin{align}
E = &-\sum_{i} \Delta E_{i}+\sum_{a}(|\partial a|-1) \Delta E_{a}, \\
\Delta E_{i}=& \frac{\beta h_{i} \sum_{x=\pm 1} x \mathcal{H}_{i}(x)+\sum_{x=\pm 1} \mathcal{G}_{i}(x)}{\sum_{x=\pm 1} \mathcal{H}_{i}(x)}, \\
\Delta E_{a}=& \beta J_{a} \frac{\tanh \beta J_{a}+\prod_{i \in \partial a} m_{i \rightarrow a}}{1+\tanh \beta J_{a} \prod_{i \in \partial a} m_{i \rightarrow a}},\\
\end{align}
\end{subequations}
where
\begin{equation}
\begin{split}
\mathcal{G}_{i}(x)&= \sum_{b \in \partial i} e^{x\beta h_{i}}\Biggl[\beta J_{b} \sinh \beta J_{b}\left(1+x \hat{m}_{b \rightarrow i}\right)\\
&+x \beta J_{b} \cosh \beta J_{b}\left(1-\tanh ^{2} \beta J_{b}\right) \prod_{j \in \partial b \backslash i} m_{j \rightarrow b}\Biggl] \\
& \times \prod_{a \in \partial i \backslash b} \cosh \beta J_{a}\left(1+x \hat{m}_{a \rightarrow i}\right).\\
\end{split}
\end{equation}
The entropy of the model is defined 
as $S=-\sum_{\boldsymbol\sigma} P(\boldsymbol{\sigma}) \ln P(\boldsymbol{\sigma})$, 
which can be also estimated from the standard thermodynamics 
relation $S = -\beta F+\beta E$.

In the entropy landscape analysis, we introduce 
a perturbed probability distribution to explore the internal 
structure of the network parameter space, which is given by
\begin{equation}
P(\mathbf{\boldsymbol{\sigma} })=\frac{1}{Z}\exp \left(\beta \sum_{i<j}{{{J}_{ij}}}{{\sigma }_{i}}{{\sigma }_{j}}+\beta\sum_{i}{{h}_{i}}{{\sigma }_{i}}+x\sum_{i}{\sigma _{i}^{*}}{{\sigma }_{i}} \right),
\end{equation}
where $\beta$ is the inverse temperature, and $x$ is the couping field tuning 
the distance between the configuration $\boldsymbol\sigma$ and the reference $\boldsymbol\sigma^*$.
With the Laplace method, the free energy density $f$ can be approximated as 
$f \approx \min _{\epsilon, q} f(\epsilon, q)$, which is given by 
$\beta f = \beta\epsilon -xq- s(\epsilon ,d)$, where $\epsilon$ is 
the energy density $(E/N)$, $q$ is the overlap $\sum_i \sigma_i \sigma_i^*/N$, $s(\epsilon,d)$ is 
the entropy density $S/N$, and $d$ is the Hamming distance per weight related to the
overlap $q$ by $d=(1-q)/2$. At the same time, $(x,q)$ should obey the following 
equations: $\partial s(\epsilon,d)/\partial d = 2x$ and $\partial s(\epsilon,d)/\partial \epsilon = \beta$, because of the Laplace method.
According to the double Legendre transform, the entropy density $s(\epsilon,d)$ is given 
by $s(\epsilon ,q)= -\beta f + \beta\epsilon -xq$. Note that, by fixing $d$ or $\epsilon$, one can construct iso-distance or iso-energy curves just through finding compatible 
$x$ or $\beta$ for the following equations:
\begin{subequations}
\begin{align}
 \frac{\partial(\beta f)}{\partial\beta}&=\epsilon,\\
 \frac{\partial(\beta f)}{\partial x}&=-q,
 \end{align}
\end{subequations}
due to the Legendre transform.

Under the perturbed probability measure, the 
cavity method is still applicable, with an only change for the bias $h_i$ as $\beta h_i\to \beta h_i + x\sigma_i^*$.
By iterating the same message passing equations to a fixed point, we can finally obtain the free 
energy density, energy density and entropy density, etc. 
Note that $q$ is computed via $\sum_i m_i \sigma^*_i/N$, where $m_i$ is the fixed-point magnetization. 

\section{Simulation details}
\label{app-c}
The MNIST handwritten digits dataset is a classification benchmark dataset which contains $60\ 000$ training images and 
$10\ 000$ test images. The image is a $28\times28$ gray-scale handwriting digit with a label taken from one of 0 to 9. For the sake of simplicity,
the dataset we use for training is re-generated by applying the PCA to the original dataset. More precisely, we apply the PCA on training 
images to compute the eigenvectors first, and then project training and test images with these eigenvectors to finally 
obtain the low dimensional (20 pixels) images. 

\begin{figure}
\centering
\includegraphics[bb=2 6 1493 545, width=.8\textwidth]{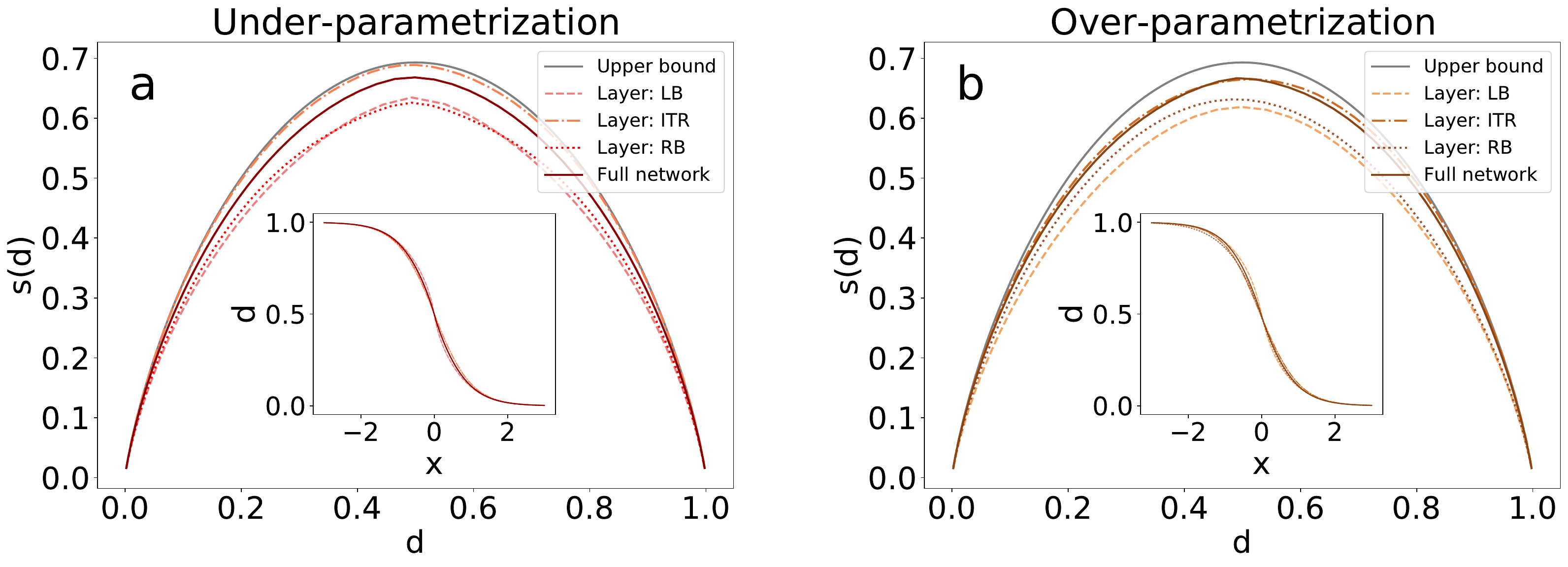}
\caption{Entropy landscape analyses in the high-energy references case. 
The reference weight configurations are randomly selected from the high-energy region, 
which displays no significant difference from the low-energy reference case [Fig.~(\ref{entropy})]. 
(a) Under-parametrization case. (b) Over-parametrization case.}
\label{S1}
\end{figure}

\begin{figure}
\centering
\includegraphics[bb=5 6 1493 545,width=.8\textwidth]{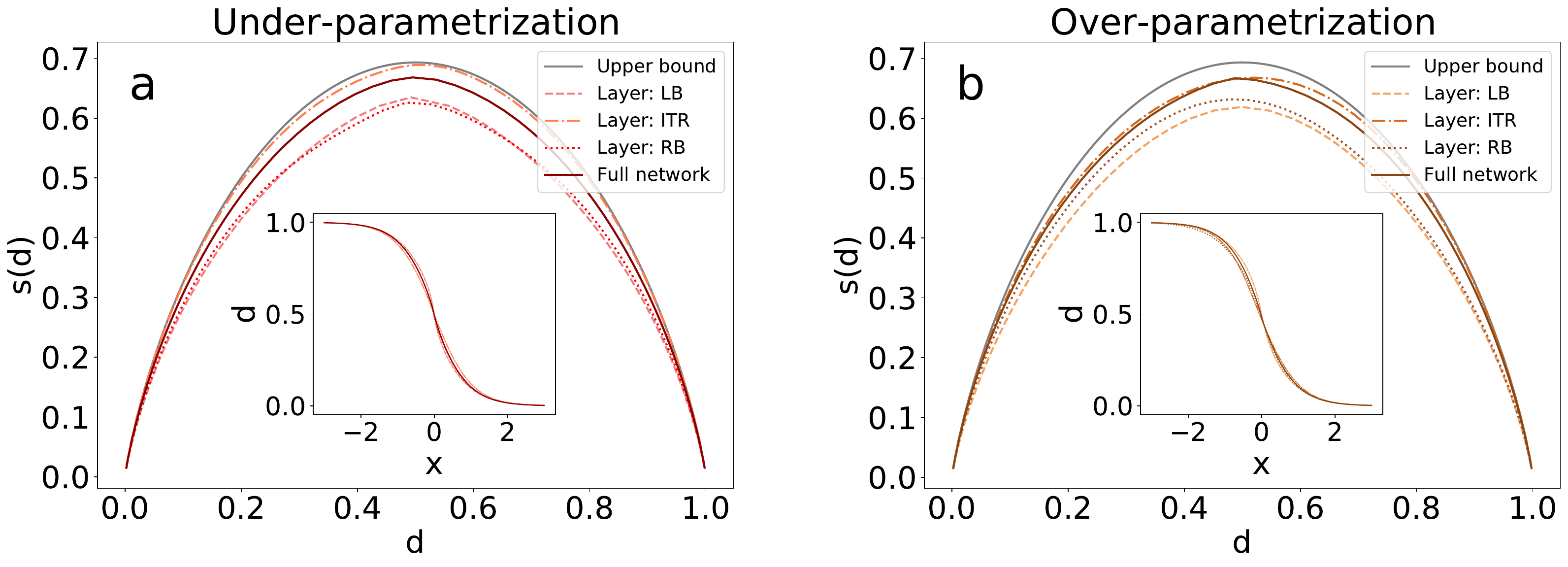}
\caption{Entropy landscape analyses in the medium-energy reference case. 
The reference weight configurations are randomly selected from the medium-energy region, 
which displays no significant difference from the low-energy reference case [Fig.~(\ref{entropy})]. 
(a) Under-parametrization case. (b) Over-parametrization case.}
\label{S2}
\end{figure}

To both have a high test accuracy and avoid a large network-parameter size at the same time, we choose 
the network architecture as 20-15-15-10, where each number denotes the corresponding layer-width.
We add a
softmax layer onto the output layer and use the cross entropy as the loss function. One training 
epoch involves the whole training dataset, which is divided into $300$ mini-batches, each of which contains $200$ images. 
The loss is optimized by Adam~\cite{Adam-2015} and the initial learning rate is set to $0.3$.
During the training, we also introduce an $L_2$ regularization to the loss function to penalize 
large external fields. The regularization strength is set to $10^{-5}$, 
which is optimal for our experiments.

The stimulation details presented above apply to the case of under-parametrization. 
We also design two other kinds of training scenarios. The first case is the over-parametrization scenario, where the 
number of training images is smaller than the number of network parameters. In this case, we only 
train $500$ images which are divided into $25$ mini-batches with the mini-batch size of $20$. Other hyper-parameters remain unchanged. The second case is the shallow-network scenario, 
where we delete all hidden layers, and thus the network architecture changes to 20-10. Other hyper-parameters 
remain unchanged as well. Following the same weight-collection procedure as the under-parametrization case except for setting 
the accuracy threshold to $70\%$, we collect one million and half a million weight configurations for the over-parameterization and shallow-network scenarios, respectively,
in order to construct effective models. 

\begin{figure}
\centering
\includegraphics[bb=5 6 1427 515,width=.7\textwidth]{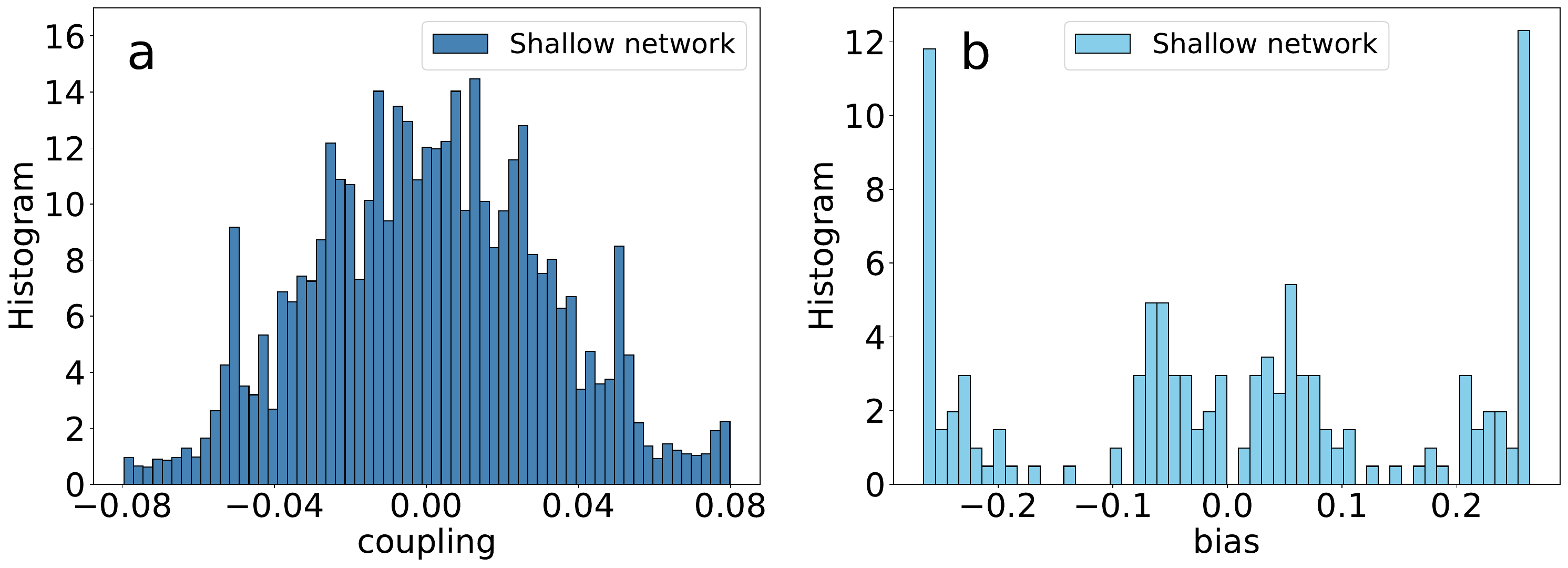}
\caption{Histograms of model parameters in the Ising model of the shallow-network case. 
(a) Histogram of inferred couplings. The couplings are distributed relatively broader compared with non-shallow networks. 
(b) Histogram of the inferred biases. The biases are also distributed broadly.}
\label{S3}
\end{figure}

\begin{figure}
\centering
\includegraphics[bb=3 4 768 547,width=.7\textwidth]{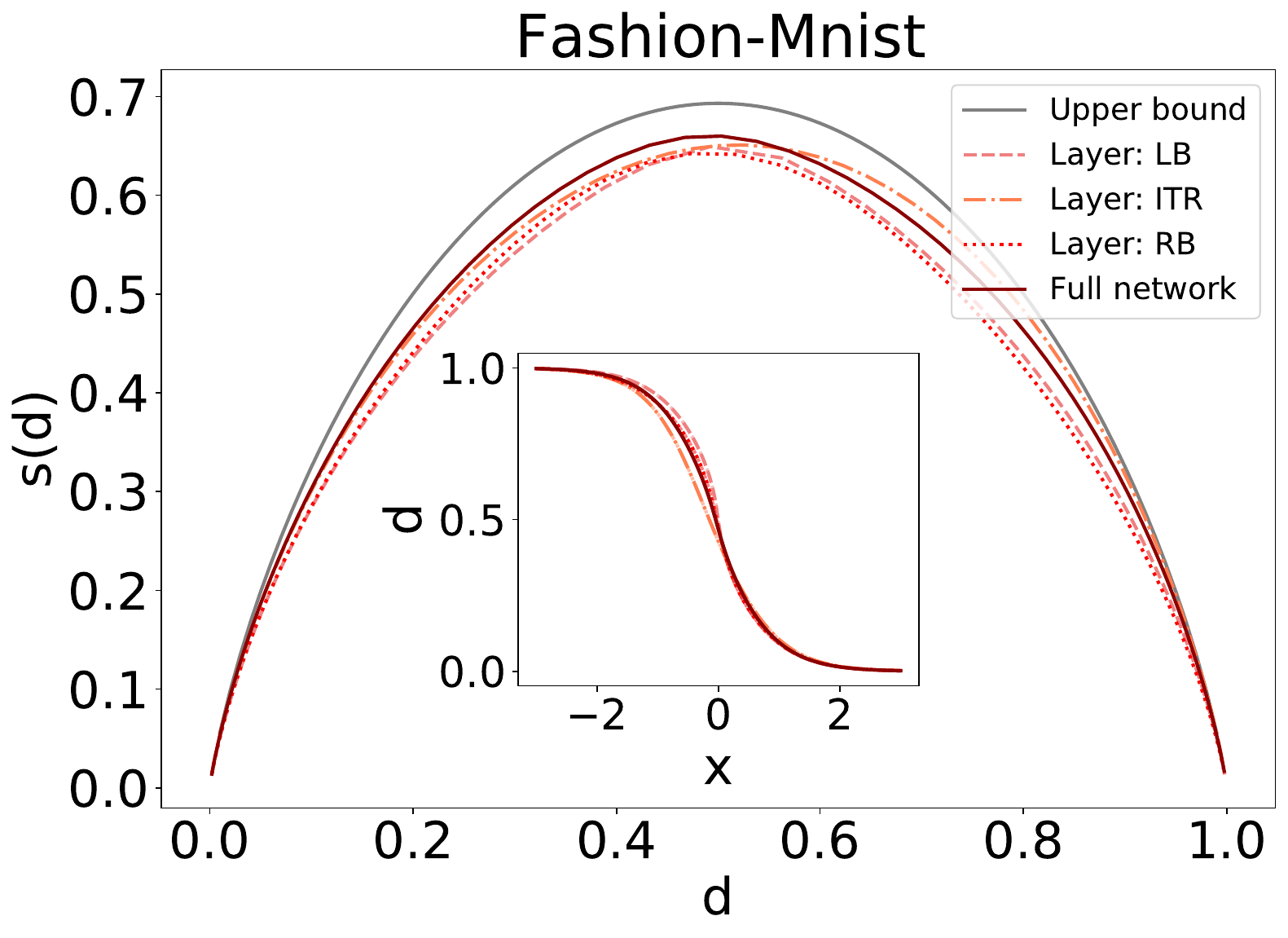}
\caption{Entropy landscape analysis on the fashion-MNIST dataset. The network structure 
remains 20-15-15-10 and the training data size is $500$ corresponding to the over-parameterization case. 
One million weight configurations after learning are collected to calculate the weight statistics, which should reach the test accuracy 
threshold equal to $70\%$. The root-mean squared errors $\Delta$ are less than $0.085$ to obtain an effective Ising model. 
To get the entropy curves, $20$ weight configurations are selected from the low energy region as the reference configurations
$\boldsymbol{\sigma}^*$. 
Other conditions are the same as in the MNIST case.
} \label{fmnist}
\end{figure}

Finally, we show additional landscape analysis when the reference weight configuration is selected from the high and medium energy regions.
A qualitative same behavior is identified (Fig.~\ref{S1} and Fig.~\ref{S2}). In addition, the histograms of model parameters inferred from the weight data
are also shown in Fig.~\ref{S3} for the shallow-network setting. Our framework can also be applied to other more challenging types of dataset, e.g., Fashion-MNIST consisting of
fashion items images belonging to ten classes~\cite{Fashion-2017}, as displayed in Fig.~\ref{fmnist} for the over-parameterization case. Unlike the case of
the MNIST dataset, the entropy landscape of the full network
gets close to that of the central part of the network, while other qualitative properties remain. 
All codes to obtain the results of this paper are available at the link: \url{ https://github.com/ZouWenXuan/Liquid-like-deep-learning}.

%%%%%%%%%%%%%%%%%%%%%%%%%%%%%%%%%%%%%%%%%%%%%%%%%%%%%%%%%%%%%55
\section{Pseudolikelihood maximization for learning Ising models}
\label{app-e}
As an alternative strategy, the pseudolikelihood method requires the entire dataset of the sampled weight configurations $\{\boldsymbol{\sigma}^\mu\}_{\mu=1}^M$, 
rather than the one- and two-point correlation functions. The objective function to be maximized is written as follows~\cite{Ising-2010},
\begin{subequations}
\begin{align}
 \label{psl}
 \mathcal{PL}&=\sum_i\mathcal{L}_i,\\
 \mathcal{L}_i&=\frac{1}{M}\sum_{\mu=1}^M\ln P(\sigma_i^\mu|\boldsymbol{\sigma}^\mu_{\backslash i}),
 \end{align}
\end{subequations}
where the conditional probability of the spin $\sigma_i$ given the rest of the system $P 
(\sigma_i^\mu|\boldsymbol{\sigma}^\mu_{\backslash i})$ is given below,
\begin{equation}
 P(\sigma_i|\boldsymbol{\sigma}_{\backslash i})=\Biggl[1+e^{-2\beta\sigma_i\Bigl(h_i+\sum_{j\neq i}J_{ij}\sigma_j\Bigr)}\Biggr]^{-1}.
\end{equation}
Therefore, the pseudolikelihood method simplifies the original $N$-body problem into $N$ independent one-body problems (but each with
$N-1$ quenched spin variables). The optimization can then be implemented through maximizing each of the $N$ local likelihood functions $\mathcal{L}_i$
separately, and finally the symmetric coupling $J_{ij}=\frac{1}{2}(J_{ij}^i+J_{ij}^j)$, where $J_{ij}^x$ is the coupling value for the maximum of
$\mathcal{L}_x$. The maximization can be done by simple gradient ascent procedures~\cite{Erik-2012}. To learn a sparse Ising model,
one can rewrite the local likelihood with $\ell_1$ regularization as $\tilde{\mathcal{L}}_i=\mathcal{L}_i+\lambda\sum_{j\neq i}|J_{ij}|$,
where $\lambda$ is a freely-tuned parameter controlling the sparsity of the model. The numerical maximization can be efficiently done by
the interior point method~\cite{Koh-2007}, where $\lambda$ is the pre-factor of $\lambda_{\rm max}$. This regularized optimization leads to a sparse effective model where non-zero and null couplings coexist, thereby
distinguishing strongly interacting weights from non-interacting ones. To save the computational cost, we fix $M=100\ 000$ in the pseudolikelihood maximization learning. 

\begin{figure}
\centering
\includegraphics[bb=8 2 1434 979,width=\textwidth]{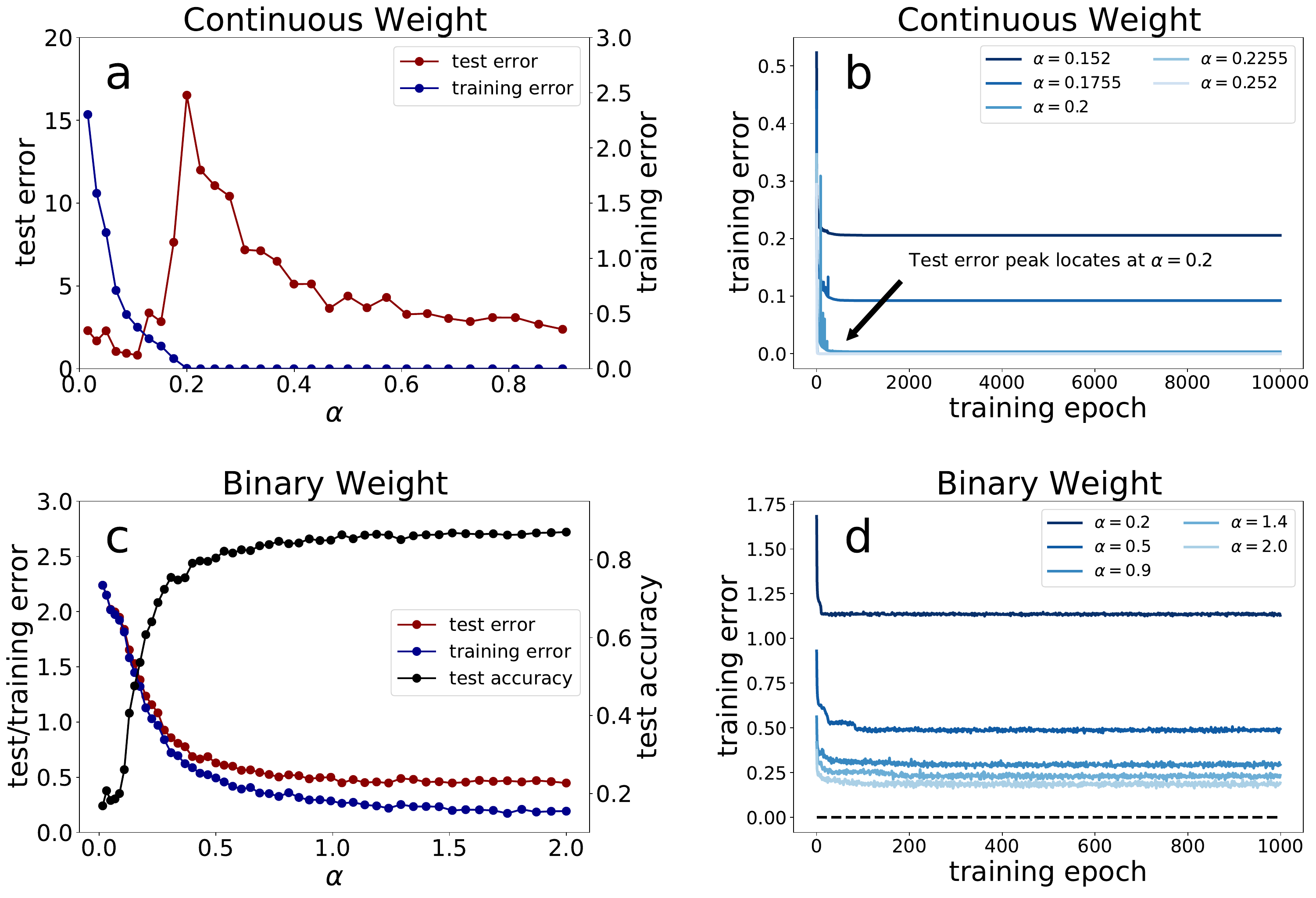}
\caption{Detection of double descent phenomenon in deep learning of continuous weights or binary weights.
The network structure is specified by 20-h-h-10, and the training data size is $2\ 000$. The ratio between the network complexity and the training data size is defined by
$\alpha$. We choose the cross entropy as the cost function.
(a,b) Deep learning of continuous weights. The point where the training error vanishes defines the interpolation point, indicating a peak in test error as well.
(c,d) Deep learning of binary weights. The double descent phenomenon is absent. We actually train the network of binary weights for a total number of $10^5$ epochs,
therefore the abscissa should be multiplied by a factor of $100$. 
} \label{ddrealbi}
\end{figure}

\section{Double descent phenomenon}
\label{app-f}
In the traditional statistical learning theory, a classical U-shaped test error curve is predicted when the training is performed with increasing model complexity.
However, deep neural networks are able to achieve the state-of-the-art performance using over-parametrized architectures.
More precisely, when the number of network parameters are sufficiently large to perfectly learn the training data (i.e., at the so-called interpolation threshold),
further increasing the model complexity will cause a second drop of the test error. Therefore, the learning behavior in the over-parametrization regime 
is called the double descent phenomenon~\cite{Belkin-2019,Spigler-2019,Hastie-2019,Mei-2019}.
Note that for ridge regression problems (implemented via a two-layer neural network for an asymptotic study,
where only the second layer weights are trained while the first layer weights are random for the sake of theoretical analysis
~\cite{Engel-2001,Belkin-2019,Hastie-2019,Mei-2019}), the test error may diverge as the interpolation boundary is approached~\cite{Hastie-2019,Mei-2019}, because of the divergence of network weights~\cite{Somp-2020}.
However, these studies are restricted to deep neural networks with real-valued weights and 
standard stochastic gradient descent training. In contrast, our study focuses on bounded synapses (e.g., $\pm1$), showing intrinsic differences.
In this section, we provide details to check if the double descent phenomenon remains in our current setting,
i.e., training deep networks of binary weights.

We consider the architecture of $K$-$h$-$h$-$x$, where $K$ means the dimension of the input image (the image can be preprocessed by PCA),
$h$ means the size of hidden layers (i.e., $N_1=N_2=h$ in the main text), and $x=2$ or $10$ means binary classification (parity of the digits, see also Ref.~\cite{Spigler-2019}) or ten-class classification respectively.
For the parity classification task, we use the hinge loss, while we use the cross-entropy loss for the ten-class classification task.
The value of $h$ is used to tune the model complexity. The value of $K$ takes from $10$ to $50$ (preservation of $82\%$ original information).
In all these settings, we found double descent phenomenon for networks of real-valued weights [see Fig.~\ref{ddrealbi} (a,b)].
In contrast, as shown in Fig.~\ref{ddrealbi} (c,d), for networks of binary weights, the test error keeps decreasing with the model complexity,
while the training error never reaches zero (or decreases extremely slowly). Compared to the real-valued-weight network,
it is very difficult to train the network to perfectly fit the target function using the low-precision binary weights, which is evident in our simulation results [Fig.~\ref{ddrealbi} (c,d)].
We thus conclude that no interpolation threshold exists in our current setting, and correspondingly the peak of test error is not observed.
This conclusion is robust for binary classification and other values of input dimensionality.

\section{Performance comparison between deep and shallow networks}
\label{app-g}
\begin{figure}
\centering
\includegraphics[bb=5 0 636 492,scale=0.5]{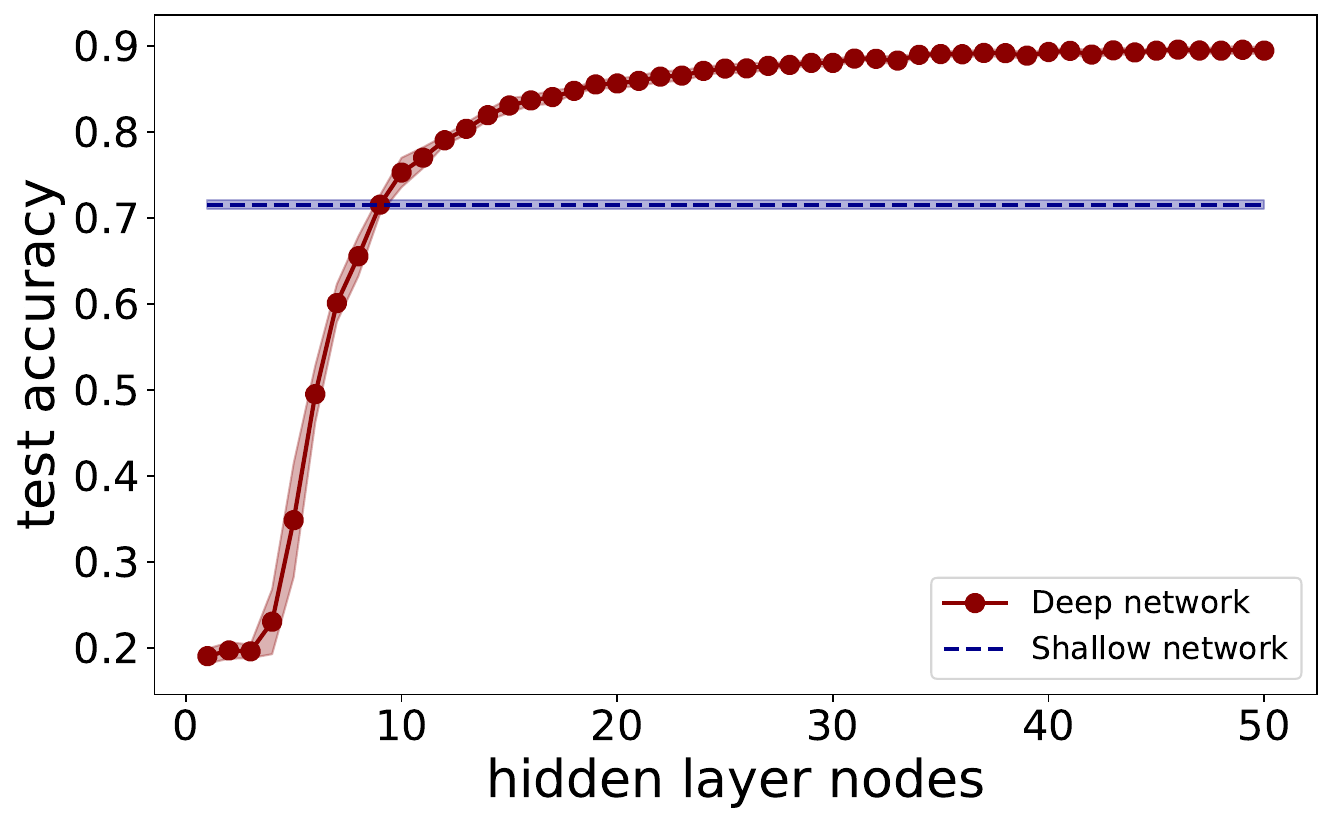}
\caption{Performance comparison between deep and shallow networks. Ten independent trainings are considered for the fluctuation.}
\label{compds}
\end{figure}

In this section, we show that the depth in our current setting can improve the performance. We consider the architecture $20$-$h$-$h$-$10$, where $h$ can be tuned.
For the shallow network, two hidden layers are removed. We train the network using the entire dataset of MNIST, and observe that
when the size of the hidden layer crosses a threshold (e.g., 10), the test performance keeps being improved over the shallow architecture.
The results are summarized in Fig.~\ref{compds}.

\section{Heat-capacity integration for entropy estimation}
\label{app-h}
When the Bethe approximation breaks down, one can use heat-capacity integration to estimate the model entropy.
According to the thermodynamics relationship, we write the entropy as follows,
\begin{equation}
 S(T=1,N)=\int_0^1\frac{C(T,N)}{T}dT,
\end{equation}
where the heat capacity $C(T,N)$ can be computed through the Monte-Carlo sampling procedure~\cite{Gasper-2009}.
More precisely, the variance of the energy, $\langle E^2\rangle-\langle E\rangle^2$, equals to $T^2C(T,N)$ (we have assumed unit Boltzmann constant).
The energy variance can be obtained by drawing a large number of samples (e.g., $2\times 10^6$) at every intermediate temperature ($T\in[0,1]$).
In fact, taking into account the coupling field term, the entropy of landscape is computed as $S-Nxq$.

\section{TAP free energy distribution}
\label{app-i}
The well-known Thouless-Anderson-Palmer (TAP) equation for a disordered Ising model reads as follows,
\begin{equation}
\label{eqtap}
m_{i}^{t+1}=\tanh \left(\beta h_i+ \beta\sum_{j(\ne i)} J_{i j} m_{j}^t-\beta^{2} \sum_{j (\neq i)} J_{ij}^{2}\left(1-(m_{j}^t)^{2}\right) m_{i}^{t-1}\right) ,
\end{equation}
where we have put the correct time indexes for iteration~\cite{IterTAP-2014}.
The fixed points of TAP are the stationary points of the TAP free energy~\cite{Georges-1991}, as given by
\begin{equation}
\begin{aligned} -\beta F_{\rm TAP} &=-\sum_{i} \left(\frac{1+m_{i}}{2} \ln \frac{1+m_{i}}{2}+\frac{1-m_{i}}{2} \ln \frac{1-m_{i}}{2}\right) \\
& +\frac{1}{2}\beta \sum_{i\ne j} J_{ij}m_i m_{j}+\frac{\beta^{2}}{4} \sum_{i\ne j} J_{ij}^{2}\left(1-m_{i}^{2}\right)\left(1-m_{j}^{2}\right).
\end{aligned}
\end{equation}
At low temperatures,
The TAP equation has many solutions with ${m_i\neq0}$, which can be interpreted as stable or metastable thermodynamic states.
Note that in our current setting, the temperature effect has been absorbed into the parameters ($\boldsymbol{\Omega}$)
of the effective Ising model. To estimate the free energy distribution, we start the iteration of Eq.~(\ref{eqtap})
from $2000$ random initializations. The results show that for deep networks, a single peak profile is observed, in contrast to the double peak profile 
for the shallow network (see Fig.~\ref{deepeak} in the main text). This observation is consistent with our entropy landscape analysis. 
%%%%%%%%%%%%%%%%%%%%%%%%%%%%%%%%%%%%%%%%%%%%%%%%%%%%%%%%%%%%%%%%%%

\begin{figure}
\centering
\includegraphics[bb=14 3 1171 572,scale=0.4]{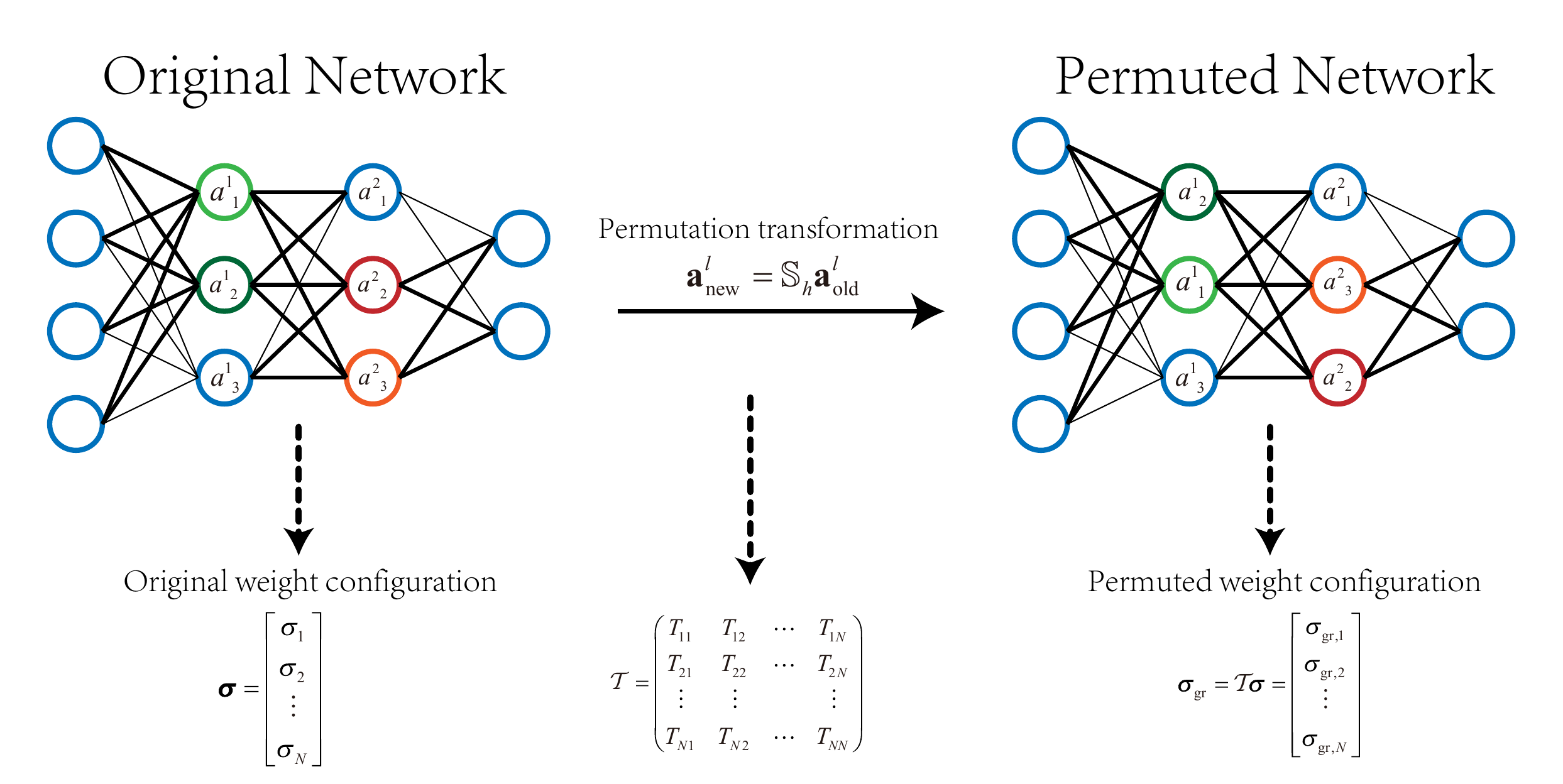}
\caption{Schematic illustration of permutation symmetry transformation. The transformation is carried out by a permutation of hidden neurons according to descending levels of pre-activation (denoted by $\mathbf{a}^l$), i.e.,
$\mathbf{a}_{\rm new}^l=\mathbb{S}_h\mathbf{a}_{\rm old}^l$, which further leads to a new weight configuration $\boldsymbol{\sigma}_{\rm gr}$, 
related to the original one $\boldsymbol{\sigma}$ via $\boldsymbol{\sigma}_{\rm gr}=\mathcal{T}\boldsymbol{\sigma}$, where the matrix entry takes either zero or one.
The thick edges represent the involved weights for the transformation, i.e., the incoming and outgoing synapses for the permuted neurons.
}
\label{psfig}
\end{figure}

\section{Node-permutation symmetry}
\label{app-j}
It is clear that there exists node-permutation symmetry for hidden layers, e.g, exchange of an arbitrary pair of nodes (including their associated incoming and outgoing weights) does not 
cause the change of the network output. This raises an interesting question---how much this permutation affects the weight configuration data and further the landscape of the effective 
model. To address this question, we first select all digits of label $1$ in the test set as the input, and then each sample of weight configuration produces
the hidden pre-activations, whose mean values over the input ensemble are subsequently ranked in the descending order for each hidden layer (called $\mathbb{H}_D^\ell$).
The ranked pre-activation result thus corresponds to
an element of the node-permutation group, which transforms the original weight configuration $\boldsymbol\sigma$ to a permuted one $\boldsymbol\sigma_{\rm gr}$ by a permutation symmetry transformation $\mathcal{T}$:
$\boldsymbol{\sigma}_{\rm gr}=\mathcal{T}\boldsymbol{\sigma}$ (see also Fig.~\ref{psfig}). Note that each element in the transformation matrix $\mathcal{T}$ takes either zero or one.
We thus call $\boldsymbol{\sigma}_{\rm gr}$ as a generating weight-configuration, from which permutation-dependent
configurations can be generated by carrying out one of the permutation transformations. The total number of these transformations is given by $h!\times h!$, where
$h$ is the width of hidden layer ($N_1=N_2=h$). We remark that the result does not depend on the selected input ensemble, e.g.,
other digits.

According to this rule, if two weight configurations lead to the same $\boldsymbol{\sigma}_{\rm gr}$ (identical $\{\mathbb{H}_D^\ell\}$ as well), we say they are permutation-dependent; otherwise, they are permutation-independent.
In fact, our original training data of weight configurations are completely permutation-independent, probably due to the limited-size weight-data yet
the extremely large size of the deep learning weight space.
To introduce the permutation symmetry among weight configurations,
we use $M_{\rm s}$ of $\boldsymbol{\sigma}_{\rm gr}$ (by a random selection) to generate as many as $2M_{\rm s}$ permutation-dependent configurations (one $\boldsymbol{\sigma}_{\rm gr}$ is paired with the corresponding generated configuration).
These generated configurations are combined with the other $M-M_{\rm s}$ ones ($M$ is the number of the collected weight configurations). Therefore, the permutation symmetry ratio is calculated as $r_{\rm ps}=\frac{2M_{\rm s}}{M-M_{\rm s}+2M_{\rm s}}$.
The effects of $r_{\rm ps}$ on the landscape are discussed in the main text.

%%%%%%%%%%%%%%%%%%%%%%%%%%%%%%%%%%%%%%%%%%%%%%%%%%%%%%%%%%%%%%%%%%%%%
%\bibliography{ref}

%%%%%%%%%%%%%%%%%%%%%%%%%%%%%%%%%%%%%%%%%%%%%%%%%%%%%%%%%%%%%%%%%%%%%

\end{document}